\documentclass[journal]{IEEEtran}
\def\@IEEEclspkgerror{\ClassError{IEEEtran}}
\usepackage{enumitem}
\usepackage{cite}
\usepackage{graphicx}
\usepackage{algorithm}
\usepackage{algorithmic}
\usepackage{amsmath}
\usepackage{amssymb}
\usepackage{amsthm}
\newtheorem{theorem}{Theorem}
\usepackage{soul}
\usepackage[dvipsnames]{xcolor}
\usepackage{hyperref}
\usepackage[dvipsnames]{xcolor}

\makeatletter
\newcounter{parenttheorem}

\makeatother

\def\BibTeX{{\rm B\kern-.05em{\sc i\kern-.025em b}\kern-.08em
    T\kern-.1667em\lower.7ex\hbox{E}\kern-.125emX}}
\usepackage{lineno}
\begin{document}
\title{Learning Vehicle Trajectory Uncertainty}

\author{Barak Or and Itzik Klein, 
    \thanks{Preprint version. 2023.}
\thanks{Authors are with the Hatter Department of Marine Technologies. Charney School of Marine Science, University of Haifa, Haifa, 3498838, Israel (e-mail: barakorr@gmail.com).}}

\markboth{Learning Vehicle Trajectory Uncertainty / Or and Klein}%
{Or and Klein: Learning Vehicle Trajectory Uncertainty}
\maketitle

\begin{abstract}
A novel approach for vehicle tracking using a hybrid adaptive Kalman filter is proposed. The filter utilizes recurrent neural networks to learn the vehicle's geometrical and kinematic features, which are then used in a supervised learning model to determine the actual process noise covariance in the Kalman framework. This approach addresses the limitations of traditional linear Kalman filters, which can suffer from degraded performance due to uncertainty in the vehicle kinematic trajectory modeling. Our method is evaluated and compared to other adaptive filters using the Oxford RobotCar dataset, and has shown to be effective in accurately determining the process noise covariance in real-time scenarios. Overall, this approach can be implemented in other estimation problems to improve performance.
\end{abstract}


\maketitle
\section{Introduction}
\label{sec:introduction}
{In many applications, accurate positioning is required. Those include}, vehicle tracking tasks \cite{bar2004estimation,kalman1960new, aloi2007comparative,xiong2020imu,pavkovic2008adaptive, qin2021improved,motroni2020sensor,xiong2022path}, vehicle trajectory smoothing \cite{baek2017accurate,xiong2020imu,marzbani2019autonomous,cui2019sigma,or2021kalman}, {autonomous driving} \cite{kala2013motion,sharma2021recent}{,  advanced driver-assistance systems} \cite{perumal2021insight} {, and 
path planning and tracking for vehicles and robots} \cite{gray2013stochastic,zhang2021enabling}.
{For trajectory uncertainty estimation, in} \cite{hu2022vehicle} {the authors present a vehicle trajectory prediction method that takes into account aleatoric uncertainty to improve the overall accuracy of the prediction. In} \cite{jiang2022vehicle}, {the authors propose a probabilistic vehicle trajectory prediction method based on a dynamic Bayesian model that integrates the driver's intention, maneuvering behavior, and vehicle dynamics using in-vehicle sensors. The method estimates the vehicle trajectory and achieves accurate long-term predictions in both lane-keeping and lane-changing scenarios.}

To achieve accurate positioning the model-based Kalman filter (KF) is widely used. One of the challenges to consider when applying a KF for tracking applications is the modeling of the vehicle trajectory, as expressed by the system matrix and associated process noise covariance. 
{Typically, constant velocity (CV) or constant acceleration (CA)} models are employed for a wide range of vehicle tracking problems \cite{or2021kalman,bar2004estimation,olama2008position,fu2020decision,stubberud2007online}. {However, these models make assumptions that may not accurately reflect real-world scenarios, where a vehicle's velocity and acceleration may vary over time. In the CV model, the underlying assumption is that the vehicle travels with constant velocity. The addition of process noise turns the model to a nearly CV model allowing the filter to cope with varying velocity conditions. In the same manner, the CA model assumes constant vehicle acceleration and the addition of process noise turns the model to a nearly CA model allowing the filter to cope with varying acceleration. That is, the added process noise covariance enables the filter to cope with unmodeled vehicle dynamics (for example, as additional states) or perturbations up to some order of magnitude }\cite{abbeel2005discriminative}.

{Using the CV and CA} models ensures a Gauss-Markov (GM) process and provides an optimal estimation \cite{zhang2020identification,jazwinski2007stochastic}. As long as the underlying model assumptions hold, the filter accuracy is satisfactory. However, in practice, the vehicle dynamics can differ from the model assumptions leading to a mismatch between the modeling and the actual behavior of the vehicle. As a consequence, the KF tracking filter performance degrades, and, in some situations, diverges. In automated vehicle perspective, it might result in a loss of information on the vehicle location that could potentially lead to an accident \cite{nilsson2016lane}. Therefore, model uncertainty, namely process noise covariance determination, is considered critical in the pre-processing phase of the KF \cite{artunedo2020motion, zhang2020identification,or2022hybrid}.

{Generally, increasing the value of the process noise covariance matrix results in  the short memory behavior of the filter. In such situations, the assumption made on the vehicle's kinematic model is less relevant, as the filter gives more weight to the external measurements when calculating the filter gain. However, in some situations, this can lead to non-optimal trajectory estimation. In contrast, setting lower values for the process noise covariance matrix leads to a long memory behavior of the filter. In such situations, the kinematic model is more dominant in the filtering process. As a result, even a small deviation from the assumed kinematic model can result in poor} estimation performance and sometimes in filter divergence.
\begin{figure}[h]
\centering
{\includegraphics[width=0.48\textwidth]{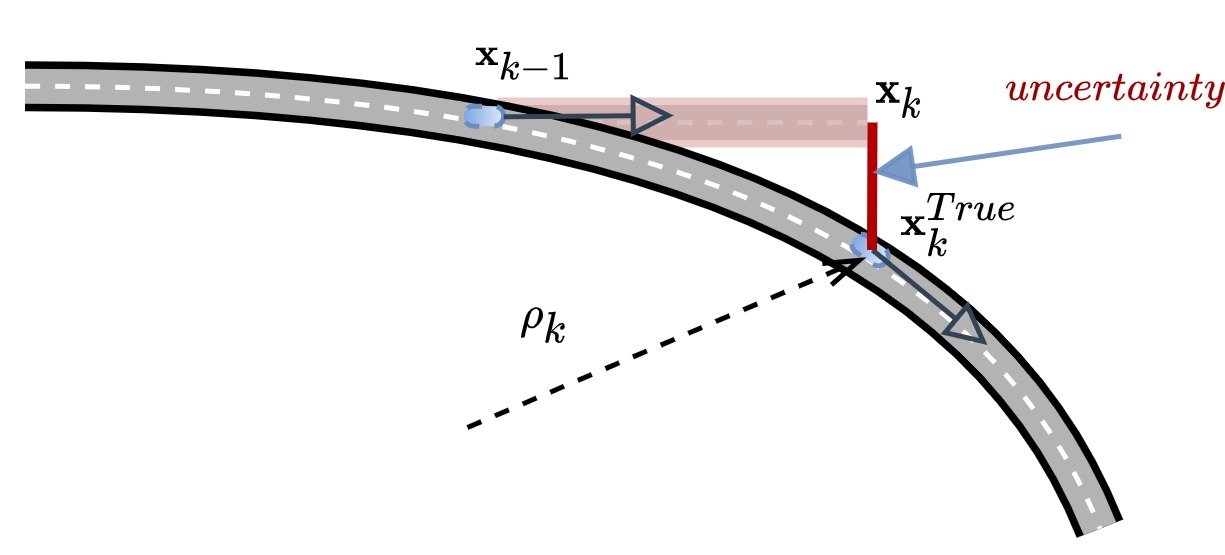}\label{fig1}}
\caption{Trajectory modeling error demonstration. The vehicle position at timestamp $k$ is predicted to maintain a straight line, according to the trajectory model (CV). However, its true position at time $k$ is along the road, where it performed a circular motion (along a circle with radius $\rho$) to {remain} on the road. {This tracking error is directly caused by uncertainty in the trajectory model design (represented by the red vertical line).}}%
\end{figure}
{To cope with varying conditions along the trajectory, model-based adaptive approaches to tune the process noise covariance, were suggested in the literature} \cite{zhang2020identification,aghili2016robust,mohamed1999adaptive,yang2006optimal,galben2011new}. Among them, the common approaches are 1) the innovation-based method, where the innovation quantity is used together with the Kalman gain to calibrate the matrix \cite{ mehra1970identification}, 2) modified scaling, where the measurement noise covariance matrix is known and used to keep the trajectory uncertainty noise matrix scaled \cite{saha2013robustness}, and 3) generative learning, where the system transition matrix model is used to calculate and correct model error every two successive steps \cite{ abbeel2005discriminative}. Although many works addressed this issue, the problem of optimal trajectory modeling by tuning the trajectory uncertainty process noise covariance is still considered unsolved \cite{zhang2020identification}. 

{Recent advancements in machine learning (ML) and deep learning (DL) techniques} \cite{goodfellow2016deep} {have demonstrated state-of-the-art performance in various fields such as computer vision} \cite{guo2016deep}, natural language processing \cite{cambria2014jumping}, inertial sensing \cite{klein2022data}, and autonomous underwater vehicle navigation \cite{cohen2022,or22022hybrid}. 

Recent works, addressing model uncertainty, explore the possibility of using ML/DL approaches. Learning approaches suggest including a loss function that contains the process noise covariance, both directly and indirectly \cite{abbeel2005discriminative}. {There,} the learning approach learns the covariance matrix parameters online. Yet, the resulting time to converge is too long, as the entire history of the tracking process is considered. In \cite{baek2017accurate}, a curve-fitting approach was proposed for tuning the trajectory uncertainty noise matrix using a neural network. {However, this method requires a training set with accurate ground truth for the entire trajectory, which makes it unrealistic for real-time tracking in new environments. This is a significant limitation of the proposed method and poses a challenge to its practical implementation in real-world scenarios.}

In \cite{or2022hybrid},{we proposed a novel hybrid learning framework for a nonlinear extended KF. This filter fuse between an inertial navigation system (INS) and a global navigation satellite system using a DL model. The DL model captures the dynamic of the system in real-time and adapts it to improve its performance. The proposed method is tested using field experiments with a quadrotor, and it was shown to improve position accuracy by 25\% compared to model-based INS/GNSS fusion.}\\

In this paper, a hybrid adaptive linear {KF}, based on {ML} algorithms and the model-based KF equations is proposed to cope with the model uncertainty in a vehicle tracking problem. {Our objective is to provide a robust estimate of the process noise covariance, resulting in enhanced positioning accuracy.  To that end, we implement the CV and CA models and assume the availability of position measurements received for example from radar, LiDAR, or GNSS receiver}. Although an end-to-end network is simpler to implement, it does not give any intuition to the problem and acts as a black box. On the other hand, hybrid approaches rely on well-established model-based theory while adding the benefits of data-driven approaches by replacing a single operation in the model-based solution. The proposed hybrid approach consists of two steps. First, recurrent neural networks {(RNNs)} \cite{sherstinsky2020fundamentals} are employed to learn the vehicle's geometrical and kinematic features{,} namely, the road curvature and vehicle speed. Secondly, these features are inserted into a supervised learning (SL) model providing the actual process noise covariance to use in the KF framework.

The main contributions of this paper are: 
\begin{enumerate}
    \item {A hybrid supervised adaptive learning-based method that incorporates road curvature, vehicle speed, and noise measurement covariance as features to accurately estimate the model trajectory uncertainty.}
    \item  {A novel neural network structure to regresses the road curvature based on past position estimates.}
    \item {Real-time implementation of a linear KF, where the learned trajectory uncertainty is utilized to optimize the model noise covariance matrix for two tracking models:  CV and CA.}
\end{enumerate}
The proposed approach can be applied in online scenarios. To demonstrate its performance, vehicle tracking using CV and CA models is considered. The proposed approach is analyzed and compared to six other adaptive filters using the Oxford RobotCar database \cite{maddern20171}. Results show the benefits of implementing the proposed {hybrid learning model} approach.

The rest of this paper is organized as follows: Section {2} deals with the problem formulation for CV and CA models and the tracking filter and its trajectory uncertainty as well as the common approaches to adapt the trajectory uncertainty noise matrix. Section {3} presents our trajectory uncertainty learning scheme {including the} feature engineering, dataset generation process, and online tuning scheme with the KF. Section {4} presents the results and Section {5} gives the conclusions.

\section{Problem formulation}
{This section presents the tracking filter by examining three different cases of trajectory uncertainty. Additionally, the CV and CA models, along with their corresponding measurement models, are provided. Finally, various model-based adaptive approaches for determining the value of the trajectory uncertainty noise matrix in a real-time setting are discussed.}
\subsection{Tracking Filter}
The linear discrete KF is presented. The filter's initial conditions  are
\begin{equation}
\begin{array}{l}
{{{\bf{\hat x}}}_0} = {\cal E}\left[ {{{\bf{x}}_0}} \right]\\
{{\bf{P}}_0} = {\cal E}\left[ {\left( {{{\bf{x}}_0} - {\cal E}\left[ {{{\bf{x}}_0}} \right]} \right){{\left( {{{\bf{x}}_0} - {\cal E}\left[ {{{\bf{x}}_0}} \right]} \right)}^T}} \right]
\end{array}
\end{equation}
where ${\bf x}_0$ is the initial state vector, ${{{\bf{\hat x}}}_0}$ is the initial estimate state vector, ${\bf P}_0$ is the initial covariance error, {${\cal E}$ is the expected value operator, and $^T$ is the transpose operator}. \\
The state estimate propagation is made by
\begin{equation}
{\bf{\hat x}}_k^ -  = {\bf{\Phi }}{{{\bf{\hat x}}}_{k - 1}}
\end{equation}
where $\hat{\bf{x}}_{k}^-$ is the estimate of the current state $k$, ${\bf \hat{x}}_{k-1}$ is the estimate of the previous state ($k-1$), and ${\bf \Phi}$ is the discrete system matrix.\\
The error covariance propagation is calculated using
\begin{equation}
{\bf{P}}_k^{\bf{ - }}{\bf{ = \Phi P}}_{k - 1}{{\bf{\Phi }}^{T}}{\bf{ + }}{{\bf{Q}}_k}
\end{equation}
where ${\bf P}_{k-1}$ is the estimate from the previous state, ($k-1$), ${\bf P}_{k}^-$ is the estimate from the current state, $k$, and $\bf Q$ is the trajectory uncertainty noise matrix (covariance). When a measurement is available the state and covariance update are made according to the following set of equations  $(4)-(7)$.\\
The Kalman gain is given by 
\begin{equation}
{{\bf{K}}_k}{\bf{ = P}}_k^ - {{\bf{H}}^T}{\left( {{\bf{HP}}_k^ - {{\bf{H}}^T} + {{\bf{R}}_k}} \right)^{ - 1}}
\end{equation}
{where $\bf R$ is the measurement noise covariance matrix and $\bf H$ is the observation matrix.}\\
The innovation vector is defined as:
\begin{equation}
{{\bf{\nu }}_k}{\bf{ = }}{{\bf{z}}_k} - {\bf{H\hat x}}_k^{\bf{ - }}
\end{equation}
{where ${\bf z}_k$ is the measurement at time step $k$.} 
{Finally,} the state estimate update is 
\begin{equation}
{{{\bf{\hat x}}}_k}{\bf{ = \hat x}}_k^ - {\bf{ + }}{{\bf{K}}_k}{\bf{\nu }}_k
\end{equation}
and the error covariance update (correction) is given by
\begin{equation}
{\bf{P}}_k {\bf{ = }}\left( {{\bf{I}} - {{\bf{K}}_k}{\bf{H}}} \right){\bf{P}}_k^ -
\end{equation}
The choice of $\bf Q$ can be divided into three categories:
\begin{enumerate}
\item {$\bf Q=0$}: might lead to a non-stabilized/optimal filter
\begin{theorem}
If ${{{\bf{Q}}_k} \to {\bf 0} }$, then the Kalman gain is a function of the weighted trajectory model, the error covariance, ${\bf P}_{k-1}$, and the measurement noise covariance only. In that manner, {Eq.(4)} can be rewritten as
\begin{equation}
{{\bf{K}}_k} = {\left[ {{{\left( {{\bf{\Phi }}{{\bf{P}}_{k - 1}}{{\bf{\Phi }}^T}} \right)}^{ - 1}} + {{\bf{H}}^T}{\bf{R}}_k^{ - 1}{\bf{H}}} \right]^{ - 1}}{{\bf{H}}^T}{\bf{R}}_k^{ - 1}
\end{equation}
\end{theorem}
\item {${\left\| {\bf{Q}} \right\|_2} \to \infty$ (flat prior)}: leads to not considering the model, so the KF uses the measurement only. 
\begin{theorem}
If ${\left\| {\bf{Q}} \right\|_2} \to \infty$, then the Kalman gain reduces to \begin{equation}
{{\bf{K}}_k} = {\left[ {{{\bf{H}}^T}{\bf{R}}_k^{ - 1}{\bf{H}}} \right]^{ - 1}}{{\bf{H}}^T}{\bf{R}}_k^{ - 1}
\end{equation}
and the state estimate Eq. $(6)$ reduces to the weighted least squares estimator (LSE):
\begin{equation}
{\bf{\hat x}}_k^{} = {\left[ {{{\bf{H}}^T}{\bf{R}}_k^{ - 1}{\bf{H}}} \right]^{ - 1}}{{\bf{H}}^T}{\bf{R}}_k^{ - 1}{{\bf{z}}_k}
\end{equation}
\end{theorem}
\item {An arbitrary positive matrix} can be set manually or using one of the  methods presented in Section {2.3}.
\end{enumerate}

\subsection{CV and CA Models}
Any trajectory model includes some level of uncertainty. The simplest modeling approach is to set a linear trajectory motion with an additional uncertainty term. This assumption has important properties such as the Markov-Gauss process, where the current state depends only on the last state, allowing optimal implementation of the KF for accurate tracking. For example, consider an autonomous vehicle tracking problem using a CV model as demonstrated in Figure 1. In practice, the vehicle moves along a circle. The velocity vector points towards the tangential direction due to the circular motion. However, according to the assumed trajectory {in the design process,}, the vehicle should be moving in a straight direction. {Such contradiction will lead to large positioning errors unless the process noise covariance is tuned online}. 

Commonly, two trajectory models are considered for vehicle motion:
\begin{enumerate}
    \item \textbf{CV model}: {Assumes that the vehicle moves at a  constant velocity in most parts of its trajectory.}
    \item \textbf{CA model}: {Assumes that the vehicle moves in a  constant acceleration in most parts of its trajectory.}
\end{enumerate}
{As the CV model is described by fewer state variables compared to CA},  it is less sensitive to modeling errors. Alongside, the CA model considers the velocity changes with an additional state for the acceleration. \\
{in the CV model, the vehicle position is modeled by}
 \begin{equation}
p_{k + 1}^i = p_k^i + v_k^i\Delta t + \sqrt {q_k^i} n_k^i
\end{equation}
{while in the CA model, the vehicle position is modeled by}
\begin{equation}
p_{k + 1}^i = p_k^i + v_k^i\Delta t + \frac{1}{2}a_k^i\Delta {t^2} + \sqrt {q_k^i} n_k^i
\end{equation}
where $p$ is the vehicle's position, ${v}$ is the vehicle's velocity, $a$ is the vehicle's acceleration, $\Delta t$ is the step size (assumed to be constant), $q$ is the uncertainty term, $n$ is a standard Gaussian white noise, and $i \in \left\{ {x,y} \right\}$ is the axis index.\\ 
The {two-dimensional} state vector of the CV model is given by
\begin{equation}
{\bf{x}} = {\left[ {\begin{array}{*{20}{c}}
{\bf{p}}^T&{\bf{v}}^T
\end{array}} \right]^T} \in {{\mathbb{R}}^{4 \times 1}}
\end{equation}
where {${{\bf{p}}^T} = \left[ {\begin{array}{*{20}{c}}
{{p^x}}&{{p^y}}
\end{array}} \right]$ and ${{\bf{v}}^T} = \left[ {\begin{array}{*{20}{c}}
{{v^x}}&{{v^y}}
\end{array}} \right]$}. \\
The CV model transition matrix is
\begin{equation}
{{\bf \Phi} ^{CV}} = \left[ {\begin{array}{*{20}{c}}
{{{\bf{I}}_{2 \times 2}}}&{\Delta t{{\bf{I}}_{2 \times 2}}}\\
{\bf{0}}&{{{\bf{I}}_{2 \times 2}}}
\end{array}} \right] \in {{\mathbb{R}}^{4 \times 4}}
\end{equation}
where ${\bf I}_{n \times n}$ is the identity matrix {of rank n}, and the process noise covariance matrix is given by
\begin{equation}
{\bf{Q}}_k^{CV} = \left[ {\begin{array}{*{20}{c}}
{{{\bf{0}}_{2 \times 2}}}&\vline& {{{\bf{0}}_{2 \times 2}}}\\
\hline
{{{\bf{0}}_{2 \times 2}}}&\vline& {\begin{array}{*{20}{c}}
{q_k^x}&0\\
0&{q_k^y}
\end{array}}
\end{array}} \right] \in {\mathbb{R}^{4 \times 4}}
\end{equation}
The {two-dimensional} state-vector of the CA model {includes the position and velocity states, as the CV model, with the addition of the acceleration state}:  
\begin{equation}
{\bf{x}} = {\left[ {\begin{array}{*{20}{c}}
{\bf{p}}^T&{\bf{v}}^T&{\bf{a}}^T
\end{array}} \right]^T} \in {{\mathbb{R}}^{6 \times 1}}   
\end{equation}
{where ${{\bf{a}}^T} = \left[ {\begin{array}{*{20}{c}}
{{a^x}}&{{a^y}}
\end{array}} \right]$ and} the corresponding transition matrix {is given by}
\begin{equation}
{{\bf \Phi}^{CA}} = \left[ {\begin{array}{*{20}{c}}
{{{\bf{I}}_{2 \times 2}}}&{\Delta t{{\bf{I}}_{2 \times 2}}}&{\frac{1}{2}\Delta {t^2}{{\bf{I}}_{2 \times 2}}}\\
{{{\bf{0}}_{2 \times 2}}}&{{{\bf{I}}_{2 \times 2}}}&{\Delta t{{\bf{I}}_{2 \times 2}}}\\
{{{\bf{0}}_{2 \times 2}}}&{{{\bf{0}}_{2 \times 2}}}&{{{\bf{I}}_{2 \times 2}}}
\end{array}} \right] \in {{\mathbb{R}}^{6 \times 6}}
\end{equation}
and the trajectory noise covariance matrix is
\begin{equation}
{\bf{Q}}_k^{CA} = \left[ {\begin{array}{*{20}{c}}
{{{\bf{0}}_{2 \times 2}}}&{{{\bf{0}}_{2 \times 2}}}&\vline& {{{\bf{0}}_{2 \times 2}}}\\
{{{\bf{0}}_{2 \times 2}}}&{{{\bf{0}}_{2 \times 2}}}&\vline& {{{\bf{0}}_{2 \times 2}}}\\
\hline
{{{\bf{0}}_{2 \times 2}}}&{{{\bf{0}}_{2 \times 2}}}&\vline& {\begin{array}{*{20}{c}}
{q_k^x}&0\\
0&{q_k^y}
\end{array}}
\end{array}} \right] \in {{\mathbb{R}}^{6 \times 6}}
\end{equation}
{The state-space model, of the CV or CA methods, is defined by:}
\begin{equation}
{{\bf{x}}_{k + 1}} = {\bf{\Phi }}{{\bf{x}}_k} + \sqrt {{{\bf{Q}}_k}} {{\bf{n}}_k}
\end{equation}
where ${\bf x}_k$ is the state vector at time step $k$ and $\bf \Phi$ is the dynamic transition matrix.\\ 
The error between the true state, ${\bf x}_{k}^{True}$, and the model state, ${\bf x}_k$, is defined by:
\begin{equation}
{{\bf{\tilde x}}_k} \buildrel \Delta \over = {{\bf{x}}_k} - {\bf x}_{k}^{True}
\end{equation}
{To estimate the vehicle state-vector, Eq.(13) for CV or Eq,(16) for CA models, external position measurements are employed. The discrete external position measurement is given by}
\begin{equation}
z_k^i = {p}_k^i + \sqrt {{r_k^i}} w_k^i,\,i \in \left\{ {x,y} \right\}
\end{equation}
where $r_k^i$ is the measurement noise covariance and $w_k^i$ is a zero-mean white Gaussian noise. \\
The measurement matrix for the CV model is expressed by:
\begin{equation}\label{eq:hcv}
{{\bf{H}}^{CV}} = \left[ {\begin{array}{*{20}{c}}
{{{\bf{I}}_{2 \times 2}}}&{{{\bf{0}}_{2 \times 2}}}
\end{array}} \right] \in {{\mathbb{R}}^{2 \times 4}}
\end{equation}
Similarly, the measurement matrix for the CA model is defined as follows:
\begin{equation}\label{eq:hca}
{{\bf{H}}^{CA}} = \left[ {\begin{array}{*{20}{c}}
{{{\bf{I}}_{2 \times 2}}}&{{{\bf{0}}_{2 \times 2}}}&{{{\bf{0}}_{2 \times 2}}}
\end{array}} \right] \in {{\mathbb{R}}^{2 \times 6}}
\end{equation}
The corresponding measurement noise covariance matrix, for both models, is given by
\begin{equation}
{\bf{R}}_k^{CV} = {\bf{R}}_k^{CA} = {\bf{R}}_k = \left[ {\begin{array}{*{20}{c}}
{r_k^x}&0\\
0&{r_k^y}
\end{array}} \right] \in {{\mathbb{R}}^{2 \times 2}}
\end{equation}
{Finally, The measurement model is defined by:}
\begin{equation}
{{\bf{z}}_k}{\bf{ = H}}{{\bf{x}}_k}{\bf{ + }}\sqrt {{{\bf{R}}_k}} {{\bf{w}}_k}
\end{equation}
where $\bf{H}$ can be the CV model Eq.\eqref{eq:hcv} or CA model Eq. \eqref{eq:hca} measurement matrix. { Notice, as this work is focused on  estimating the model uncertainty (process noise covariance), without the loss of generality, the measurement noise covariance Eq.(24) is assumed to be perfectly known.} \\
Also, as there is no correlation between the measurement and process noises, the cross covariance matrix {(${\cal E}\left[ {{{\bf{n}}_k}{{\bf{w}}_k}^T} \right]$)} is assumed to be zero.
\subsection{Model-Based Adaptive Approaches}
{Three commonly used model-based approaches for adaptive tuning of the process noise covariance are addressed. Those methods are described below:} 
\begin{enumerate}
\item \textbf{Innovation-based method}. The most common approach to estimate $\bf Q$ in an adaptive KF framework, was suggested in \cite{mehra1970identification}. This approach is based on the innovation vector {(Eq.(5))} to construct the innovation matrix for a window size $\xi$ 
\begin{equation}\label{eq:invmat}
{{\bf{C}}_k} \buildrel \Delta \over = \frac{1}{\xi }\sum\limits_{j = k - \xi  + 1}^k {{{\bf{\nu }}_j}{{\bf{\nu }}_j}^T}
\end{equation}
{where ${\bf{C}_k}$ is the innovation matrix}.
The choice of $\xi$ is critical for the system performance: if a small window size is selected, the averaging might be {insufficient to capture the relevant information}. Yet, if $\xi$ is too big, there will be a delay in the innovation estimation due to the time elapsed from the previous trajectory points. \\
The innovation matrix, Eq.\eqref{eq:invmat}, together with the Kalman gain, $\bf K$, are used to adapt $\bf Q$ using:
\begin{equation}
{\bf{\hat Q}}_k = {{\bf{K}}_k}{{\bf{C}}_k}{\bf{K}}_k^T
\end{equation}
\item \textbf{Generative learning}. A naive approach for learning the filter parameters (and specifically $\bf Q$) \cite{abbeel2005discriminative} requires access to the full state-vector, and is generally considered inapplicable (due to difficulty in measuring all state variables). Generative learning demands the maximization of the likelihood function of all the data, using:  
\begin{equation}
\begin{array}{l}
{{\bf{Q}}^*} = \mathop {\arg \max }\limits_{\bf{Q}} \\
\left\{ {\begin{array}{*{20}{l}}
{ - M\log \left[ {2\pi {\bf{Q}}} \right]}\\
{ - \sum\limits_{k = 1}^M {{{\left( {{{\bf{x}}_k} - {\bf{\Phi }}{{\bf{x}}_{k - 1}}} \right)}^T}{{\bf{Q}}^{ - 1}}\left( {{{\bf{x}}_k} - {\bf{\Phi }}{{\bf{x}}_{k - 1}}} \right)} }
\end{array}} \right\}
\end{array}
\end{equation}
{where $M$ is the amount of samples.}
The optimal solution of {Eq.(28)} is given by
\begin{equation}
{{\bf{Q_k^*}}} = \frac{1}{M}\sum\limits_{k = 1}^M {\left( {{{\bf{x}}_k} - {\bf{\Phi }}{{\bf{x}}_{k - 1}}} \right){{\left( {{{\bf{x}}_k} - {\bf{\Phi }}{{\bf{x}}_{k - 1}}} \right)}^T}}
\end{equation}
\item \textbf{Scaling method}. A scaling method for the covariance matrices was proposed in \cite{ding2007improving} and \cite{hu2003adaptive}. The Kalman gain, $\bf K$, depends on the ratio of $\bf Q$ and $\bf R$. Hence, if one of them is known, the other can be estimated. For example, if $\bf R$ is known, $\bf Q$ can be estimated by satisfying the following condition for covariance matching:
\begin{equation}
{\bf{HP}}_k^ - {{\bf{H}}^T}{\bf{ + }}{{\bf{R}}_k} = {{\bf{C}}_k}
\end{equation}
The core idea behind the scaling method is that if the estimated covariance of $\nu_k$ in {Eq.(26)} is much larger than the theoretical covariance, then $\bf Q$ should be increased and vice versa. This deviation from the optimal value is considered using a scaling factor ${\bf \alpha}_k$, defined as
\begin{equation}
{{\bf{\alpha }}_k} = \frac{{trace\left( {\left[ {\frac{1}{\xi }\sum\limits_{j = k - \xi  + 1}^k {{{\bf{\nu }}_j}{{\bf{\nu }}_j}^T} } \right] - {{\bf{R}}_k}} \right)}}{{trace\left( {{\bf{HP}}_k^ - {{\bf{H}}^T}} \right)}}
\end{equation}
leading to a scaled covariance of 
\begin{equation}
{{{\bf{\hat Q}}}_k} = {\sqrt {{\bf{\alpha }}_k}}{{{\bf{\hat Q}}}_{k - 1}}
\end{equation}
\end{enumerate}
\section{Learning Trajectory Uncertainty }
{In this section, the motivation for  the importance of the trajectory uncertainty matrix is described. Next, a detailed description of our proposed hybrid approach, including  the learning methodology and the process of generating the datasets used in this study is given.}
\subsection{Motivation}
Assume that the vehicle trajectory can be divided into a finite set of curves, ${\bf L}_k \in \cal L$. For each curve, the averaged curvature, averaged velocity, and measurement noise can be calculated. The curve can be represented by the vehicle estimated states:
\begin{equation}
{{\bf{L}}_k} = \left\{ {\hat p_i^x,\hat p_i^y} \right\}_{i = k - N}^{k-1} \in {{\mathbb{R}}^{2 \times N}}
\end{equation}
where ${\bf L}_k$ is a state segment, represented by $N$ points. 
Using the curvature operator, {Eq.(34)}, the curve, as a function of the vehicle states, is mapped into {a} segment curvature, {${\bar \kappa }_k$}:
\begin{equation}
{{\bar \kappa }_k}\left( {{{\bf{L}}_k}} \right) = \frac{1}{N}\left[ {\sum\limits_{j = k}^{k + N - 1} {{\kappa _j}} } \right]
\end{equation}
where {${ \kappa }_j$} is defined by
\begin{equation}
{\kappa _j} \buildrel \Delta \over = \frac{{\det \left( {{{{\bf{L'}}}_j},{{{\bf{L''}}}_j}} \right)}}{{{{\left\| {{{\bf{L}}_j}} \right\|}^3}}}
\end{equation}
and {${\bf{L'}}_k$ and ${\bf{L''}}_k$ are the first and second order derivatives, respectively}. 
From an estimation point of view, such formulation of a curve, ${\bf L}_k$, can be interpreted as a road section $k$ provided by $N$ points with curvature $\kappa_k$, as demonstrated in Figure 2. \\
\begin{figure}[h]
\centering
{\includegraphics[width=0.48\textwidth]{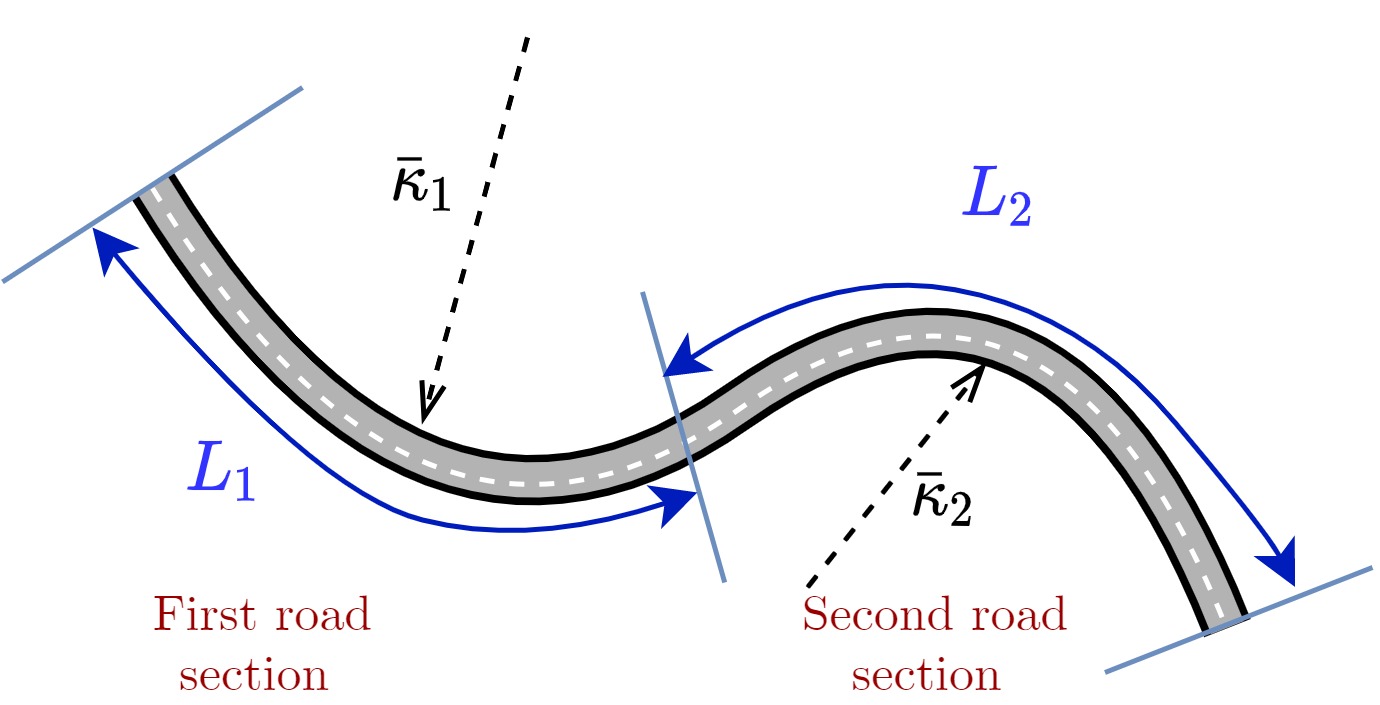}\label{fig4}
\caption{A general trajectory consisting of two sections with different geometrical properties showing the length and curvature of each segment. The road curvature is defined in Eq.(35).}}
\end{figure}
Obviously, if $\kappa  \to \infty$, the trajectory is a straight line and no compensation for the geometrical mismatch between the trajectory model and the actual vehicle trajectory should be applied. However, if the mismatch is significant 
, compensation for the inaccuracy of the assumed model should be applied{. Hence,} $\bf Q$ should be tuned accordingly. \\
{For example,} a demonstration of the a vehicle trajectory that was set to follow a circular trajectory while the KF system model was set to follow a straight line with $\bf Q=0$ is presented in Figures 3. Notice, that the KF prediction of the position vector suffered from a large error.

\begin{figure}[h]
\centering
{\includegraphics[width=0.48\textwidth]{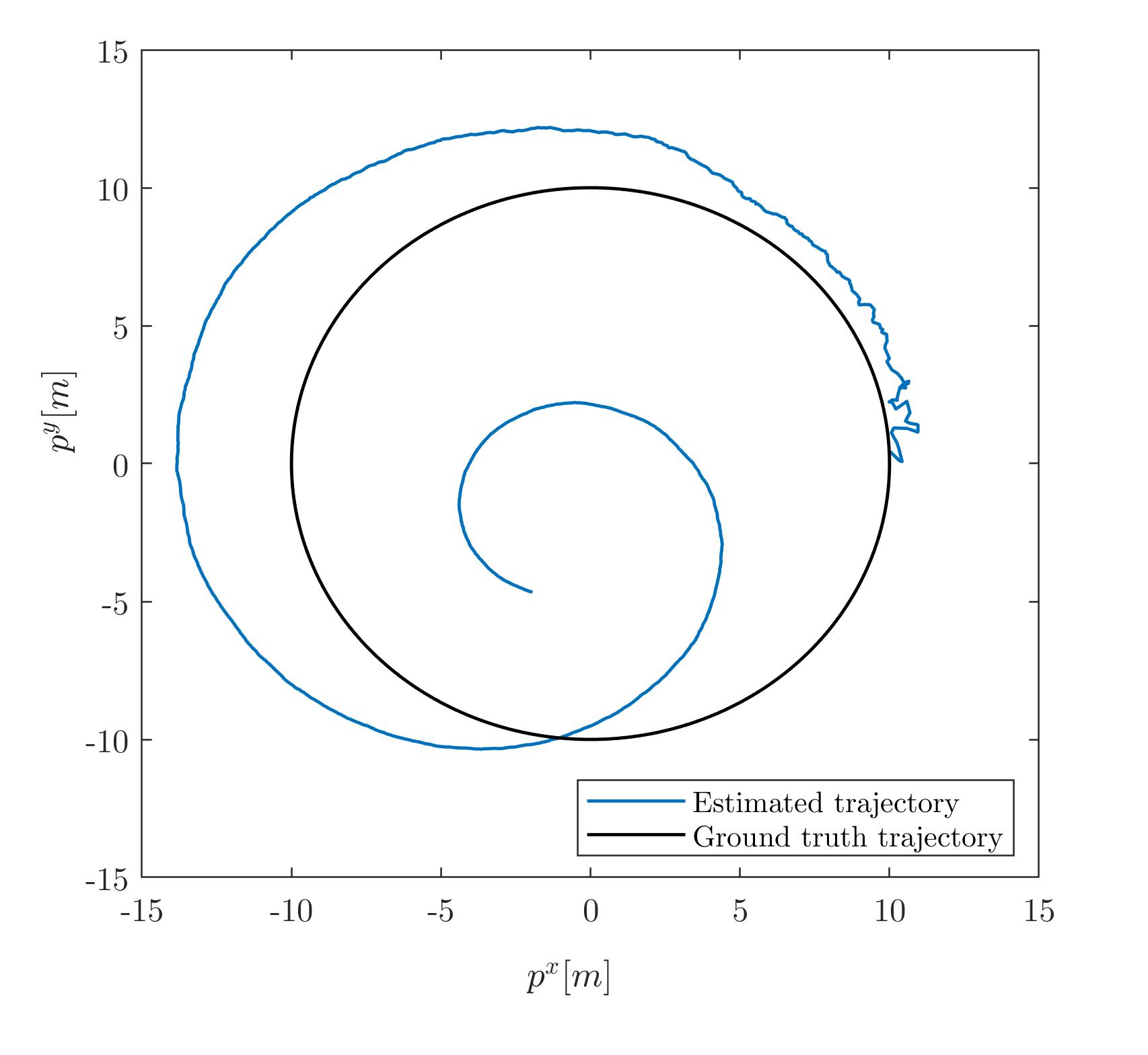}\label{fig3}}
\caption{Estimated and ground truth trajectories. The {KF} was designed with an incorrect trajectory model (the vehicle was moving in a different trajectory) and with $\bf Q=0$.}
\end{figure}
{Thus, accurate vehicle estimation requires considering various factors such as the curvature of the vehicle's path, its speed,  measurements noise, and other factors. By modeling the vehicle's trajectory as a CV or CA model, the filter designer can adjust the trajectory process noise matrix to account for these factors and their level of uncertainty. This approach can effectively handle the complexities of real-world vehicle position estimation by incorporating important features (factors) into the trajectory modeling without the need for pre-mapping road segments or using a map. This can be done in real-time, as described in the next section.} 
\subsection{Proposed Approach}
A novel approach is derived for online learning the trajectory model uncertainty, in terms of the process noise covariance matrix, using geometric and kinematic features. It is argued that an online calculation of such features together with a pre-trained model allows better determination of the trajectory uncertainty model, and hence, the process noise covariance matrix. In the proposed approach, presented in Figure 4, the previous $N$ vehicle position states are inserted into a bi-directional {long short-term memory (LSTM)} network to regress curvature of each trajectory segment. {In parallel, the vehicle} speed is calculated using the averaged estimated velocity components of the previous $N$ states. {Next, the road curvature, vehicle speed, and the measurement noise covariance, are employed as features (input)} to an additional learning model (fine tree, {Gaussian Process Regression (GPR)}, or support vector machine) to determine ${\bf Q}_k^*$. Finally, {the model-based KF utilizes the adaptive learned} optimal ${\bf Q}_k^*$ to output the current estimated state, $\hat{ \bf x}_k$.
\begin{figure}[h]
\centering
{\includegraphics[width=0.48\textwidth]{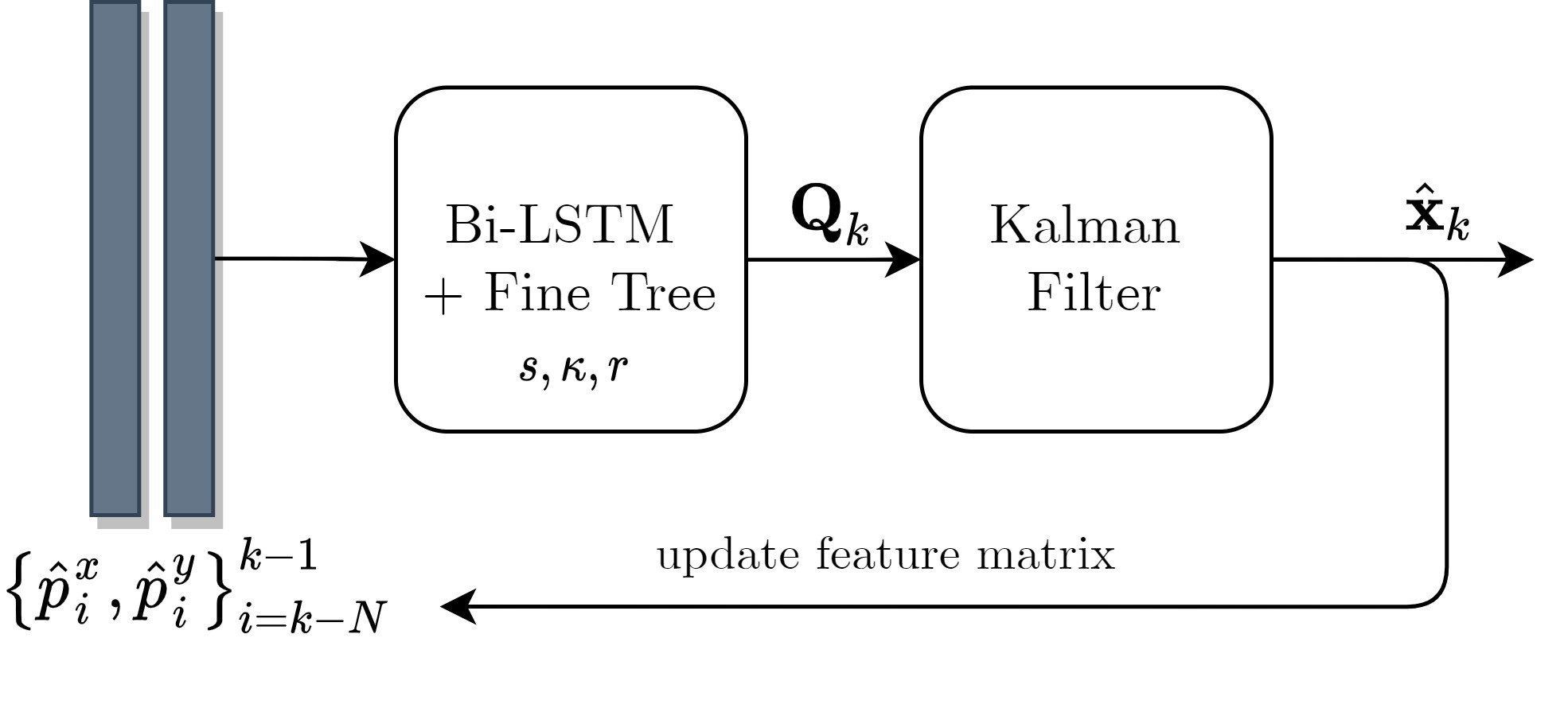}\label{fig5}
\caption{block diagram of the proposed approach. $N$ previous estimated position states are inserted into a bi-directional LSTM {to regress the road curvature}. The road curvature, vehicle speed, and the measurement noise covariance, are employed as features (input) to an additional learning model to  determine ${\bf Q}_k^*$.}}
\end{figure}
Estimating the curvature, $\hat \kappa$, in real-time vehicle tracking scenarios is not a trivial task, as the measured vehicle position and their estimates are noisy. Hence, an LSTM network was suggested to estimate the averaged curvature along a curve \cite{gers2000learning}; specifically, a bi-directional LSTM {(Bi-LSTM)}, as described below. The curvature alone is not sufficient for tuning ${\bf Q}_k$; therefore, two additional features are considered: 1) the measurement noise covariance and 2) the vehicle speed. These three features are combined into an ML model to determine the optimal ${\bf Q}_k$ according to the last $N$ measurements and their estimate. 
\subsection{Learning Framework}
The proposed learning framework has two steps:
1) The Bi-LSTM model used for online curvature learning 2) A feature-based {SL} model to estimate the process noise covariance. Here, the curvature from step 1) is used as one of the features. The proposed learning framework is illustrated in Figure 5. 
\begin{figure*}[h]
\centering
{\includegraphics[width=1\textwidth]{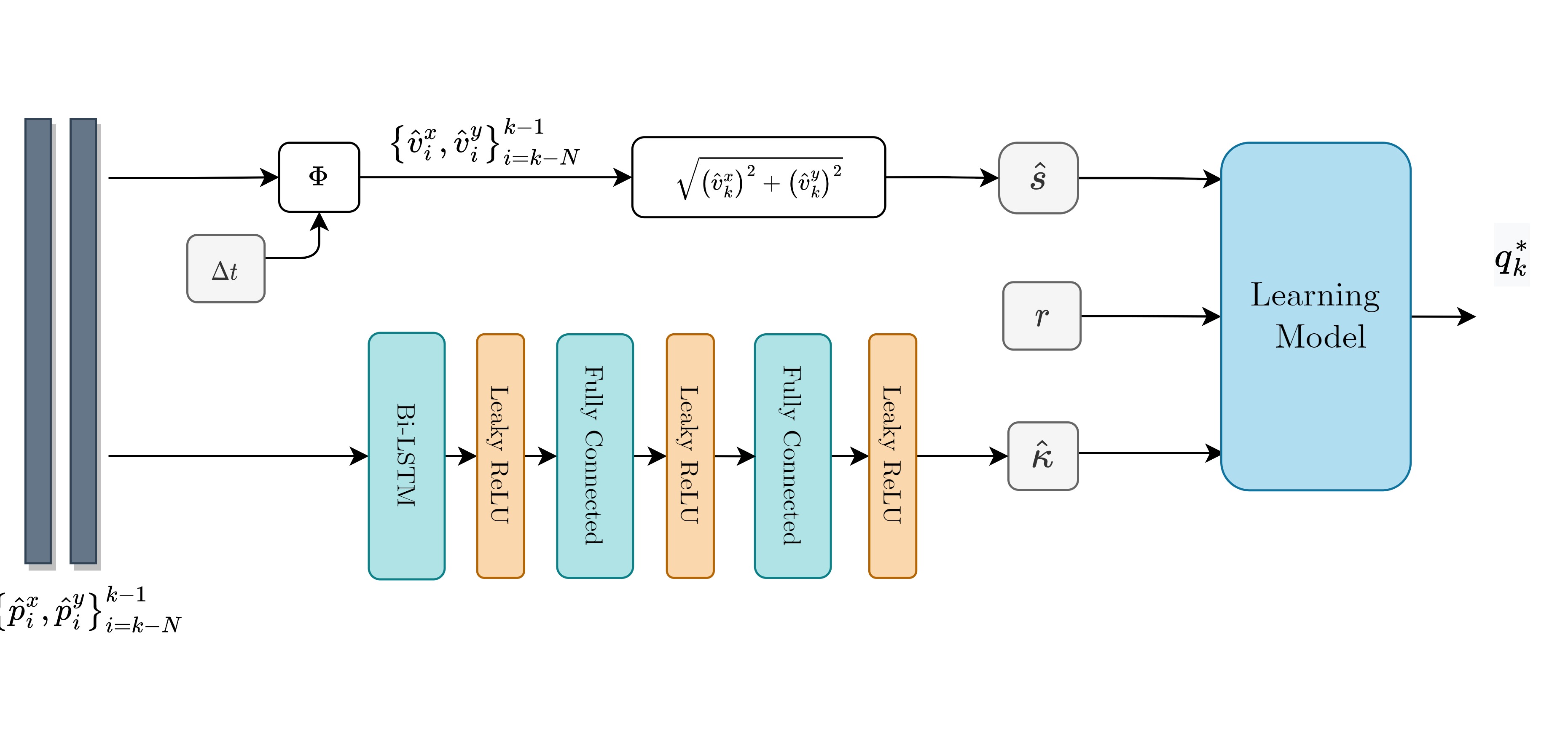}\label{fig_Net}}
\caption{The proposed learning model consists of two stages: 1) a Bi-LSTM  model to regress the road curvature ($\kappa$) and 2) the three features speed (s), noise magnitude (r), and curvature, are plugged into a learning algorithm to regress the process noise covariance}
\end{figure*}
\subsubsection{Bi-LSTM based model for online curvature learning}
A DL-based model to generalize the geometric curvature property was designed. The main advantage of using DL is the generalization capability of intrinsic properties appearing in sequential data. The LSTM is a modification of {the vanilla RNN}, where feedback connections are added. Every LSTM unit with input  (${\alpha}_t$) contains a cell ($C_t$), an input gate ($i_t$), an output gate ($o_t$), and a forget gate ($f_t$). The cell's output ($h_t$) is usually flattened and inserted into fully connected layers. The cell remembers values over intervals and the three gates regulate the flow of information in and out of the cell, as shown in Figure {6}. \\
\begin{figure}[h]
\centering
{\includegraphics[width=0.48\textwidth]{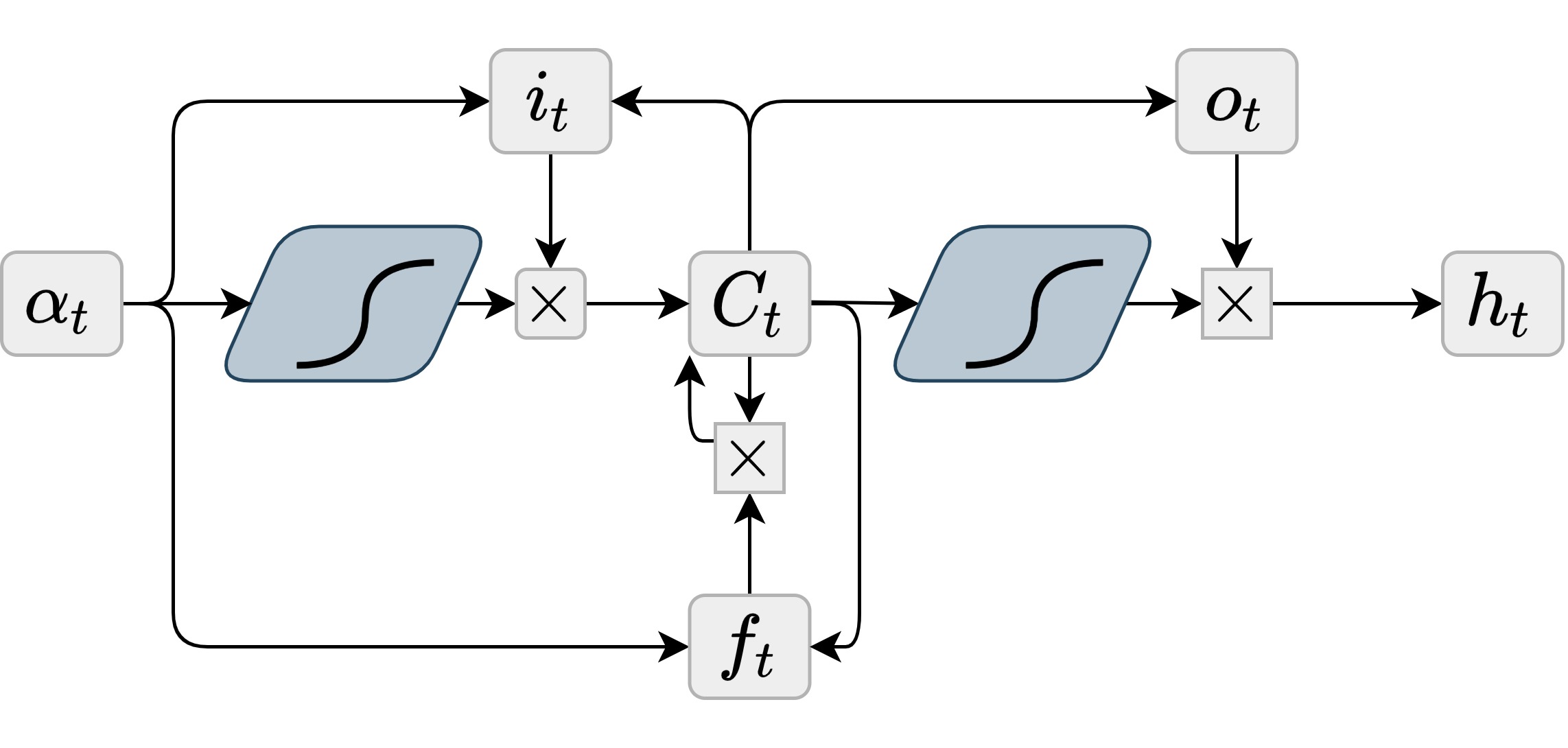}\label{fig_LSTM_block}}
\caption{LSTM unit with input, forget, and output gates.}
\end{figure}
Generally, when using an LSTM network, the information passes only backward, where only casual time-dependent properties are calculated (only past information). One method to circumvent this is to pass the information forward through an additional layer, thereby allowing the network to capture additional information. This approach is known as the bi-directional LSTM (Bi-LSTM) \cite{gers2000learning} and is illustrated in Figure {7}. Here, two time blocks are presented for two successive time steps, each with two LSTM blocks consisting of $100$ units. One passes information forward and the other passes information backward. Their output is combined using a Sigmoid activation function to obtain their final output ($m$), inserted into a fully connected layer with $30$ hidden units.
\begin{figure}[h]
\centering
{\includegraphics[width=0.48\textwidth]{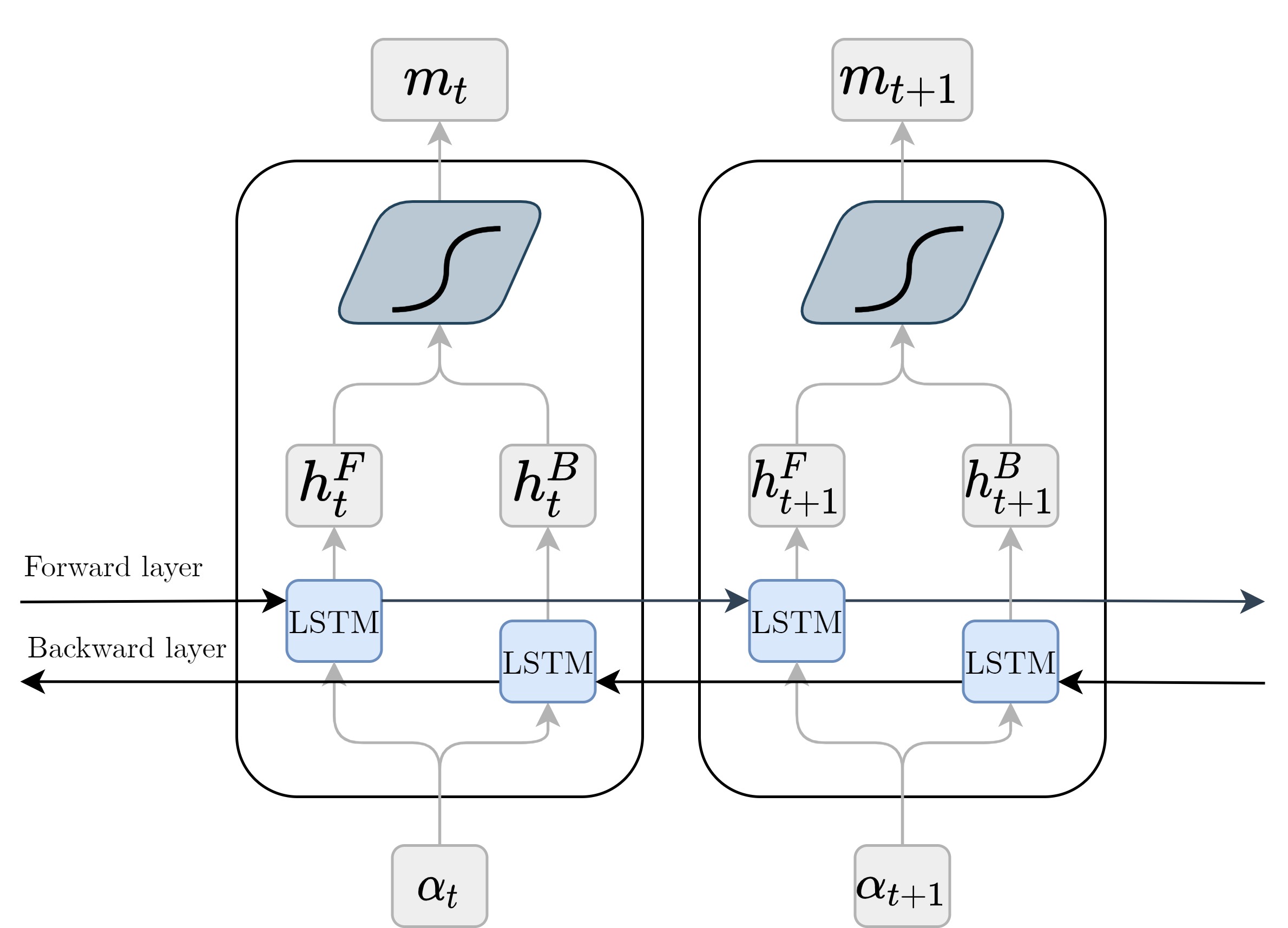}\label{fig7}}
\caption{Illustration of a bi-directional LSTM model. The output of the forward layer ($h_t^F$) is combined with the output of the backward layer ($h_t^B$) using a Sigmoid activation function.}
\end{figure}
{An example on how to produce an adaptive learning curvature module is provided in Figure {8. The figure shows} a circle with a predefined radius, equals to $10[m]$ (black line). The measurement noise magnitude was set to $1[m^2]$ in both $x$ and $y$ directions, and the vehicle speed was set to $10[m/s]$. Applying the KF {(Eq.(2)-(7))}, created the estimated trajectory (blue). Each $20$ samples along the circle were stored with their respective optimal $\kappa^*$, as the ground truth label. By repeating this procedure a dataset was created.} 

\begin{figure}[h]
\centering
{\includegraphics[width=0.48\textwidth]{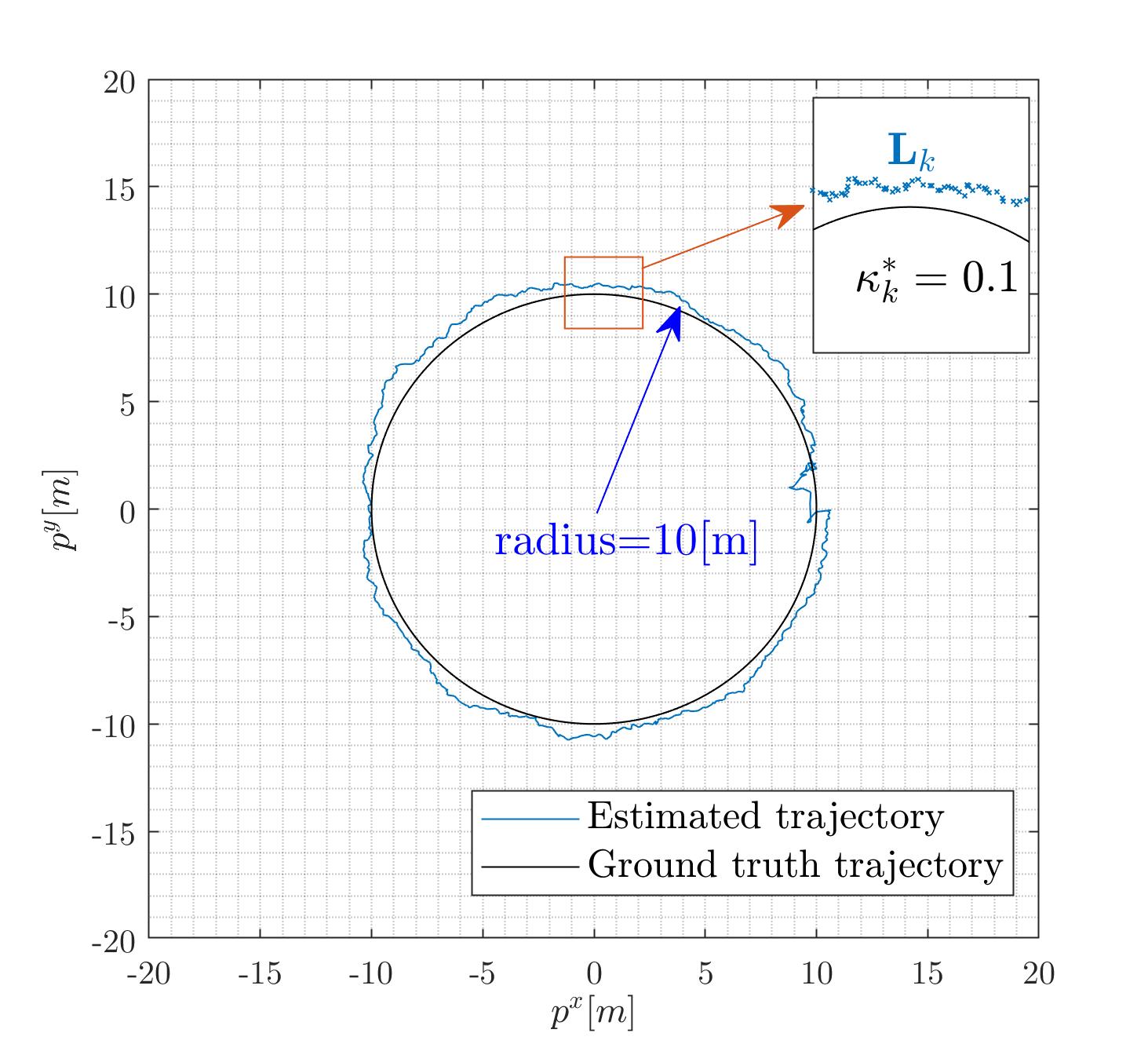}\label{fig77}}
\caption{{An example of a trajectory included in the dataset.}}
\end{figure}

\subsubsection{Learning models configuration}
Three types of features are used as input to the SL model:
\begin{enumerate}
\item Road curvature (as described in the previous subsection):
\begin{equation}
{\hat \kappa }_k = {f_{Bi-LSTM}}\left( {\left\{ {{{\bf x}_j}} \right\}_{j = k - N}^{k-1}} \right)
\end{equation}
where $\kappa_k$ is the road curvature at time $k$.
\item Vehicle speed:
\begin{equation}
{{\hat s}_k} = \sqrt {{{\left( {\hat v_k^x} \right)}^2} + {{\left( {\hat v_k^y} \right)}^2}}
\end{equation}
where $\hat{s}_k$ is the estimated vehicle speed { calculated using the estimated velocity components} at time $k$.
\item Measurement noise covariance:
\begin{equation}
{\bf{R}}_k = \left[ {\begin{array}{*{20}{c}}
r&0\\
0&r
\end{array}} \right]
\end{equation}
where $r$ is a {know} constant {(for the entire scenario)} noise variance .
\end{enumerate}
The SL model's output is $q^*$: 
\begin{equation}
q_k^* = {f_{SL}}\left( {{\hat \kappa },r,\hat s} \right)
\end{equation}
{where $f_{SL}$ is the applied learning algorithm.} 
\subsubsection{Learning algorithm summary}
Algorithm 1 summarizes the online tuning process of ${\bf Q}_k$, using both the Bi-LSTM network and the SL model, {embedded in the model-based KF framework}. In step 5, ${\bf Q}_k$ is calculated using the vehicle speed, the measurement noise covariance, and the curvature estimation. An hedging mechanism enforces ${\bf Q}_k$ as a positive diagonal matrix. Finally, the KF is updated and  the states are stored. Step 5 expresses the proposed learning framework including the Bi-LSTM{-based model and the SL model. Figure 9 shows Algorithm 1, steps 2-5,} including the network structure. {The general flow of the algorithm is summarized in Figure 5.}

\begin{algorithm}[H]
 \caption{{SL-based model for online tuning of ${\bf Q}_k$}}
 \begin{algorithmic}[1]
 \renewcommand{\algorithmicrequire}{\textbf{Input:}}
 \renewcommand{\algorithmicensure}{\textbf{Output:}}
 \REQUIRE ${\bf x}_0, {\bf P}_0,{{\bf Q}_0,{\bf R},{\bf \Phi},{\bf H},{\bf z}},\xi$ 
 \ENSURE  ${\bf \hat x}_k$
 \STATE Initialization: ${\bf \hat x}_0={\bf 0}, {\bf {\hat Q}}_1={\bf {Q}}_0$
 \\ \textit{LOOP Process}
   \FOR {$k = 1$ to  ${M}$}
   \STATE Propagate\\ ${\bf \hat x}_k^-, {\bf P}_k^-=f({\bf \Phi},{\bf \hat x}_{k-1}, {\bf P}_{k-1}^-,{\bf {\hat Q}}_k)$
   \STATE Calculate gain\\
  ${\bf K}_k=f({\bf P}^-_k,{\bf H},{\bf R})$
   \STATE Estimating speed \\
   ${{\hat s}_k} = \sqrt {{{\left( {{\hat v}_k^x} \right)}^2} + {{\left( {{\hat v}_k^y} \right)}^2}}$

   \IF {$k>\xi$}
  \STATE Update ${\bf {\hat Q}}_k$ \\ 
  ${{{\bf{\hat Q}}}_k} = {f_{SL}}\left( {s,r,\left\{ {{{\bf{x}}_j}} \right\}_{j = k - \xi  + 1}^k} \right)$
  \STATE Hedging $\bf Q$ \\
   ${{{\bf{\hat Q}}}_k} \leftarrow {{{\bf{\hat Q}}}_k} \odot {\bf{I}}$ \\
 ${\left[ {{Q_{ij}}} \right]_k} \leftarrow {\left[ {{Q_{ij}}} \right]_k}{{\cal I}_{\left\{ {{{\left[ {{Q_{ij}}} \right]}_k} > 0} \right\}}}$
    \ELSE
    \STATE ${\bf Q}_k$=${\bf Q}_0$
  \ENDIF
  
  \STATE Update \\
  ${\bf \hat x}_k, {\bf P}_k^+=f({\bf \hat x}^-_k,{\bf K}_k,{\bf z}_k,{\bf H},{\bf \hat x}_{k-1}, {\bf P}_{k-1}^-,{\bf {\hat Q}}_k)$
  \STATE Return ${\bf \hat x}_k$
 
  \ENDFOR
 \end{algorithmic}
 \end{algorithm}
\begin{figure}[h]
\centering
{\includegraphics[width=0.35\textwidth]{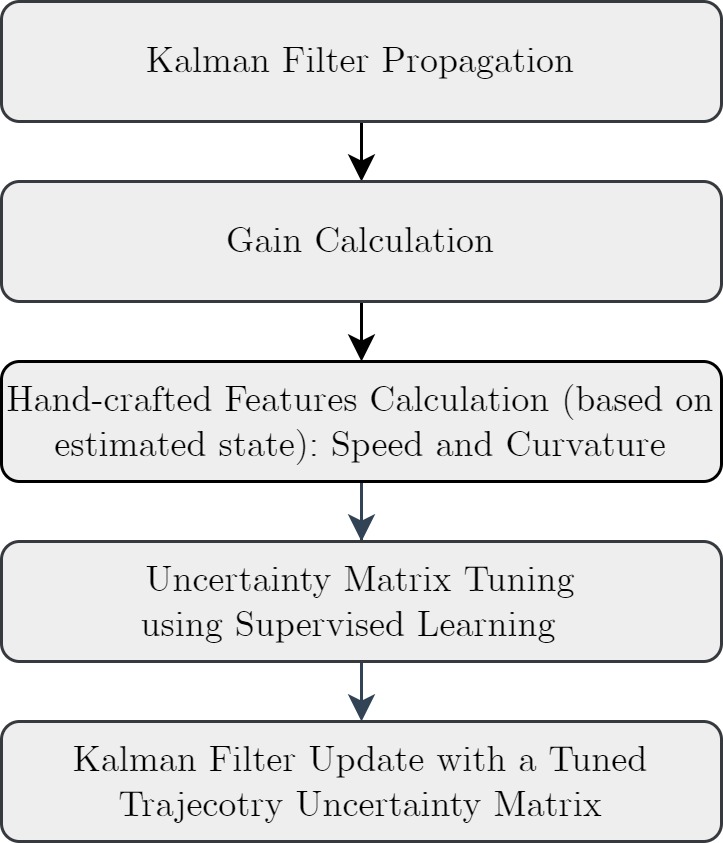}\label{fig15}}
\caption{{The proposed algorithm follows a set of steps to estimate the trajectory uncertainty matrix. First, the state vector is propagated by using the dynamic model and the previous trajectory uncertainty matrix. Then, the Kalman gain is computed, the state vector is updated, and handcrafted features are estimated based on the state vector. Finally, using those features an SL-based model is implemented to estimate the trajectory uncertainty matrix.}}
\end{figure}
%
\subsection{Datasets}
To establish the relationship between vehicle speed, measurement noise, road curvature, and their optimal ${\bf Q}^*$ matrix, the following assumptions were made:
\begin{enumerate}
\item Fixed altitude: All vehicle's trajectories are represnted in two dimensions.
\item All vehicle's trajectories can be represented as a superposition of curves, each with corresponding curvatures $\kappa_k$.
\item Vehicle speed values vary between $2[m/s]$ and $40[m/s]$. 
\item Measurement noise variance is identical in both directions such that $r_x=r_y=r$, and noise is assumed to be zero mean white Gaussian noise. 
\item The KF step-size is constant.  
\end{enumerate}
Following these assumptions, we constructed circular trajectories with different radius lengths ($1/\kappa$), measurement noise, and speed, as provided in Table {1}. The circles were obtained using the following circular trajectory model:
\begin{equation}
p_k^x = (1/\kappa) \cos \left( {\omega {t_k}} \right)
\end{equation}
\begin{equation}
    p_k^y = (1/\kappa)\sin \left( {\omega {t_k}} \right)
\end{equation}
where $t_k$ is the time propagated by 
\begin{equation}
{t_k} = {t_{k - 1}} + \Delta t
\end{equation}
and the angular velocity, $\omega$, is defined by the road's curvature, $\kappa$, and the vehicle's speed, $s$, as 
\begin{equation}
    \omega  = s \kappa.
\end{equation}
{Eqs.}(40)-(43) combined with the KF {(Eqs. (1)-(7))} were simulated once for the CV model {(Eq.(14),(15),(22),(24))} and once for the CA model {(Eq.(17),(18),(23),(24))}. 

Using the parameters in Table 1, a grid search for the optimal $q^*$ value was performed. For example, the set ${\left\{ {\kappa ,r,s} \right\}_j}$ was run 34 times for different $q$ candidates (the list is provided in the appendix). For each candidate, $q_j$, {Monte-Carlo} (MC) with $50$ iterations was simulated to obtain an accurate calculation of the {mean square error (MSE) measure}. 

Finally, the value that {minimizes the MSE} was stored with its set values as the ground truth values:
\begin{equation}
    {{\bf{Q}}^*} = \mathop {\arg \min }\limits_{\bf Q} {{\tilde x}^T}{\bf{Q}}\tilde x
\end{equation}
where
\begin{equation}
\begin{array}{l}
{\bf{Q}} = \left[ {\begin{array}{*{20}{c}}
{{{\bf{0}}_{(m - d) \times (m - d)}}}&{{{\bf{0}}_{(m - d) \times d}}}\\
{{{\bf{0}}_{d \times (m - d)}}}&{q{{\bf{I}}_{d \times d}}}
\end{array}} \right] \in {{ \mathbb{R}}^{m \times m}},
\end{array}
\end{equation}
$m$ is the state vector dimension, and $d$ is the physical dimension. Thus in our 2D scenarios, $d=2$. For example, one case of finding the optimal $q^*$ values for $\kappa=0.05[1/m], s=10[m/s], r=1[m^2]$ in a CV trajectory model is presented in Figure {9}. The optimal value obtained a position root mean square error {Eq.(46)} of less than $0.4[m]$ with $q^*=0.04$. 

\begin{table}[ht!]
\caption {{Feature parameters used to create the train dataset.}} \label{tab:title1} 
\begin{center}
\begin{tabular}{ |c|c|c| } 
\hline
Parameter & Values Range & Value step\\
\hline
Speed   & $2 < s < 40\,\,\left[ {\frac{m}{s}} \right]$ & $\Delta s = 2\left[ {\frac{m}{s}} \right]$         \\ 
Curvature     &$1/200 < \kappa  < 1{\mkern 1mu} \left[ {{\mkern 1mu} \frac{1}{m}} \right]$& $\Delta \kappa  = 1/10\left[ {\frac{1}{m}} \right]$   \\ 
Measurement noise cov.  & $0.2 < r < 4\,\,\left[ {{m^2}} \right]$&$\Delta r = 0.2\left[ {{m^2}} \right]$ \\
\hline
\end{tabular}
\end{center}
\end{table}
\begin{figure}[h]
\centering
{\includegraphics[width=0.48\textwidth]{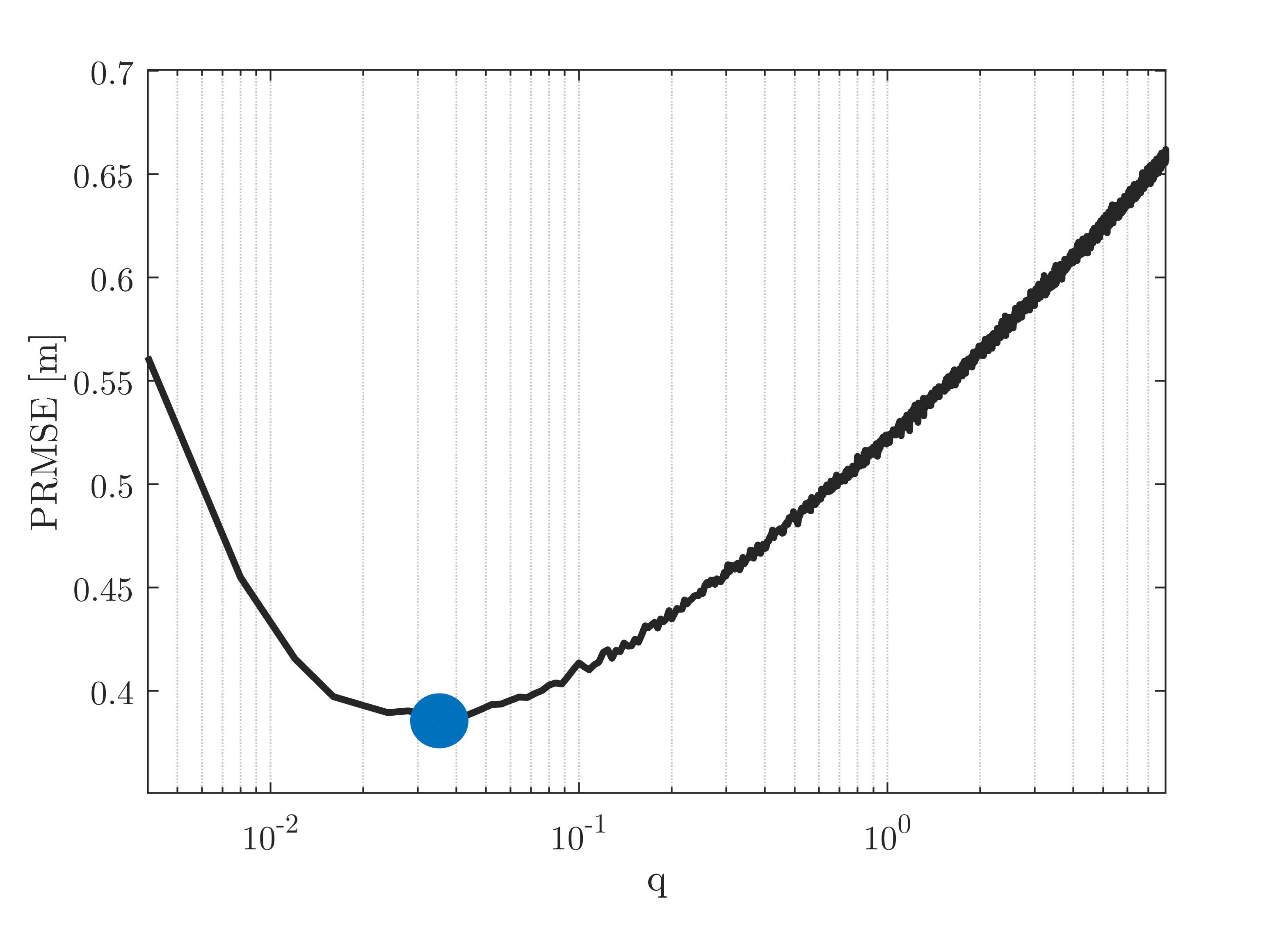}\label{fig9}}
\caption{An example of finding the optimal $q^*$ in a sense of minimum PRMSE {(Eq.(46))}.}
\end{figure}
\section{Analysis and Results}
{In this section, we present a comparative evaluation of model-based adaptive approaches and our suggested learning-based models. To assess the performance of the models, we utilize two error metrics as evaluation criteria:}
\begin{itemize}
\item {Position root mean square error is}
\begin{equation}
PRMSE = \sqrt {\frac{1}{{{T_k}}}\sum\limits_{i \in \left\{ {x,y} \right\}} {\left[ {\sum\limits_{k = 1}^{{T_k}} {{{\left( {p{{_k^i}^{True}} - \hat p_k^i} \right)}^2}} } \right]}}
\end{equation}
where $T_k$ is the total time steps.
\item {Position mean absolute error:}
\begin{equation}
PMAE = \frac{1}{{{T_k}}}\sum\limits_{i \in \left\{ {x,y} \right\}}^{} {\left[ {\sum\limits_{k = 1}^{{T_k}} {\left| {{{\left( {p_k^i} \right)}^{True}} - \hat p_k^i} \right|} } \right]} {\mkern 1mu} {\mkern 1mu}
\end{equation}
\end{itemize}
{A single RobotCar trajectory of 31 minutes/10[km], with varying vehicle driving scenarios, is employed as the test trajectory. This trajectory is illustrated in Figure 11}.
\begin{figure}
\centering
{\includegraphics[width=0.48\textwidth]{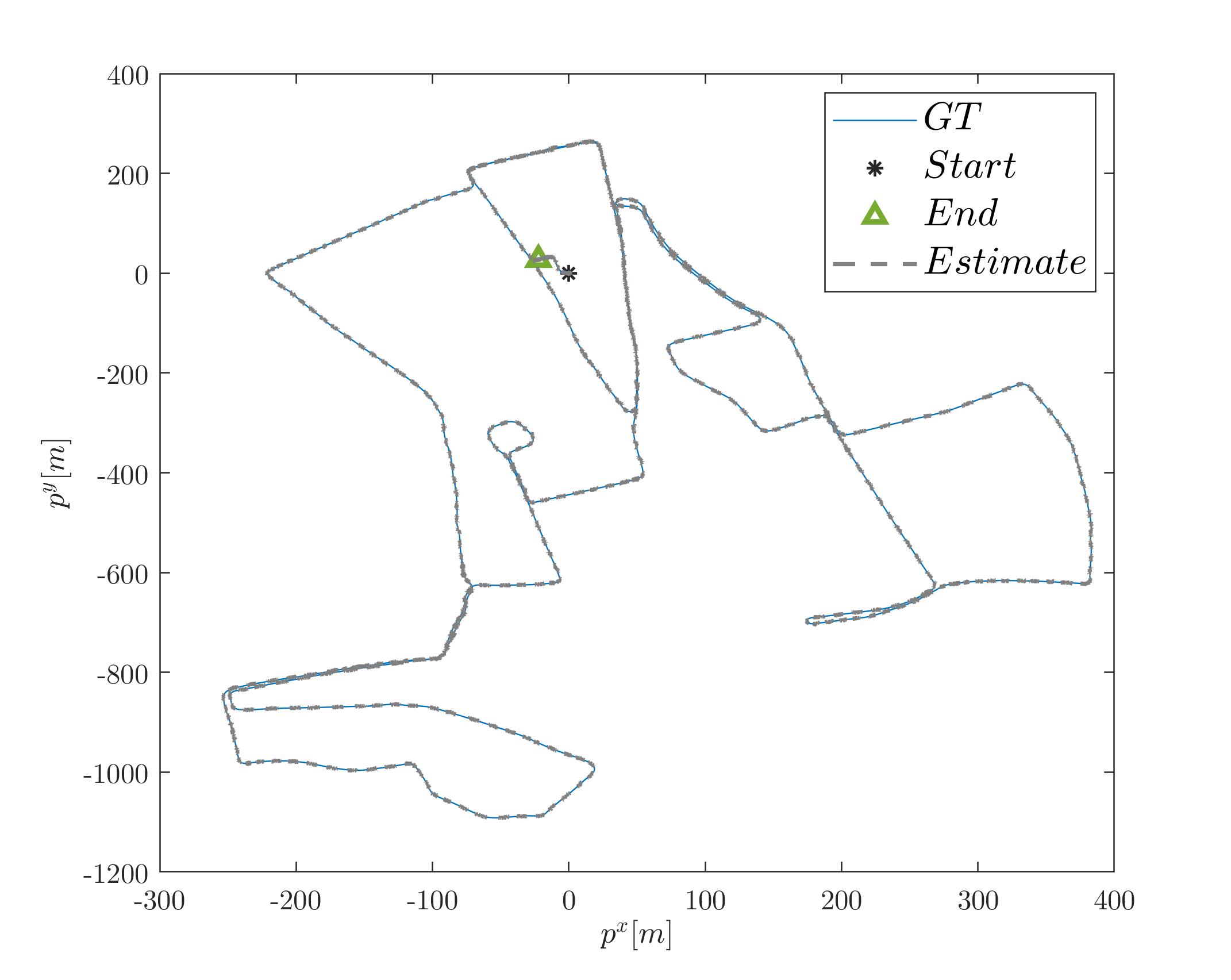}\label{fig10}}
\caption{{A single trajectory from the Oxford RobotCar dataset with a duration of $31$ minutes and $10$ kilometers of vehicle driving scenarios. The estimated trajectory was obtained using the CA fine tree Bi-LSTM based $Q(r,s,\kappa)$ Regression model, where $r=0.5[m^2]$.}}
\end{figure}
\subsection{{Road Curvature Estimation}}
In the first step of the proposed learning framework, Section {3.3}, the goal is to reconstruct the road curvature given a short curve ($N=20$). To that end, another massive dataset of $240,000$ examples was created. Each example contains twenty points (features), simulated along a defined circle with a known radius and its curvature (label). These examples were defined according to Table {1}. \\
This dataset was divided in a ratio of $80/20$ for the train/test procedures ($192,000/48,000$), where for every epoch, $24,000$ shuffled examples were fed {into the LSTM-based model} with their respective curvature parameters. {Our working environment includes: Intel i7-6700HQ CPU@2.6GHz 16GB RAM with MATLAB. The training time of the final curvature estimator elapsed about 20 minutes.} The Adam optimizer \cite{kingma2014adam} with a gradient threshold, for $30$ epochs (8 iterations each) was employed for the training. In this setup, the trained model achieved a curvature root mean square error {(RMSE)} of {0.0114}$[m^{-1}]$ on the test set.\\
{Six architectures for the curvature estimation task were suggested and trained:}
\begin{enumerate}
\item {\textbf{ BiLSTM+FC}: A BiLSTM layer with 20 hidden units, followed by a fully connected (FC) layer with ten units and output layer of a unit representing the curvature size.}
\item {\textbf{ BiLSTM+2FC}: Same as BiLSTM+FC  with an additional FC layer with ten units.}
\item {\textbf{ BiLSTM+2FC+LReLU}: Same as BiLSTM+2FC  with a nonlinear activation function between the layers, leaky rectified linear unit (LReLU), to deal with the non-linearity property of the learning task.}
\item {\textbf{ LSTM+FC }: One LSTM layer with 20 units followed by one FC layer. }
\item {\textbf{ LSTM+2FC }: Same as LSTM+FC with an additional FC layer with 10 units.}
\item {\textbf{ LSTM+2FC+LReLU }: Same as LSTM+2FC with the LReLU activation functions between the layers.}
\end{enumerate} 
{The RMSE of each model and the average running time (mean value of 1000 runs) are provided in Table 2. The best architecture, BiLSTM+2FC+LReLU, achieved a curvature RMSE of 0.0114$[m^{-1}]$ on the test set.}
\begin{table*}[h!]
\caption {{DL-based models performance in estimating the road curvature.}} \label{tab:title2} 
\begin{center}
\begin{tabular}{ |c|c|c| } 
\hline
Architecture & RMSE $[m^{-1}]$ &running time [s] \\
\hline
BiLSTM+FC &0.0460 & 0.0042  \\ 
\hline
BiLSTM+2FC&0.0240 & 0.0042  \\ 
\hline
BiLSTM+2FC+LReLU (chosen) & {\bf 0.0114} & 0.0044 \\ 
\hline
LSTM+FC &0.0257 & 0.0039 \\ 
\hline
LSTM+2FC &0.0401 &0.0042  \\ 
\hline
LSTM+2FC+LReLU &0.0131 & 0.0042 \\ 
\hline
\end{tabular}
\end{center}
\end{table*}
\subsection{{Model-Based Approaches}}
The {model-based adaptive approaches}, presented in Section {2.3}, were evaluated using Monte Carlo (MC) simulations with 100 iterations on the RobotCar \cite{maddern20171} dataset (test dataset), for various measurement noise levels as summarized in Table {3}. 
First,  the cases with $q \to 0$ (Theorem 1), which were translated for practical reasons into $q=10^{-9}$ to avoid numerical issues, were evaluated. This case represents the high confidence level of the KF in the system modeling.  There, as expected, {significant position errors were} obtained for all noise levels for both CV and CA models, . Next, scenarios with  $q \to \infty$ (Theorem 2), which were translated into $q=10^{9}$ to avoid numerical issues, were examined. This case represents a flat prior, where the trajectory model is not considered, and the position error is mainly determined by measurement. The chosen constant values were the same as the measurement noise covariances: $q=r$. {As expected, the RMSE was greatly improved compared to the case of} $q \to 0$. {Then, the adaptive model-based methods were evaluated.} The innovation-based $\bf Q$ tuning approach, generative learning (with four/six terms along the diagonal of $\bf Q$), and scaling method were implemented with a fixed window size of $\xi=10$ as no major change in the performance for different window sizes was observed. All adaptive approaches yield better performance than  $q \to 0$,  $q \to \infty$, or constant values. {In most cases, the} lower PRMSE and PMAE were obtained for the innovation-based method with a CA model. 
\begin{table*}[h!]
\caption {Comparison of six model-based approaches on the test dataset with different values of the measurement noise covariance. Each approach was examined on the CV and CA models.}
\label{tab:title3} 

\begin{center}
\begin{tabular}{ |c|c|c|c|c|c|c| } 
\hline
Model / Error metric & PRMSE [m] & PMAE [m] & PRMSE [m] & PMAE [m] & PRMSE [m] & PMAE [m] \\
\hline
Measurement noise cov. & $r=0.5[m^2]$ & $r=0.5[m^2]$ & $r=2[m^2]$ & $r=2[m^2]$ & $r=4[m^2]$ & $r=4[m^2]$ \\
\hline
CV ${\bf Q} \to 0$ &$98.4$ & $103$ &$125$ &$132$ &$143$ &$151$\\
CA  ${\bf Q} \to 0$ &$23.6$ &$24.5$ &$32.0$ &$33.0$ &$37.5$ &$38.4$\\
\hline
CV ${\left\| {\bf{Q}} \right\|_2} \to \infty$ (LSE) &$0.99$ &$1.12$ &$1.99$ &$2.25$ &$2.82$ &$3.19$ \\
CA ${\left\| {\bf{Q}} \right\|_2} \to \infty$ (LSE) &$1.00$ &$1.12$ &$2.00$ &$2.25$ &$2.82$ &$3.18$\\
\hline
CV constant $\bf Q$ &$0.64$ &$0.73$ &$1.12$ &$1.26$ &$1.47$ &$1.66$\\
CA constant $\bf Q$ &$0.62$ &$0.70$ &$1.13$ &$1.27$ &$1.53$ &$1.72$\\
\hline
CV innovation-based $\bf Q$ &$0.64$ &$0.70$ &$1.09$ &$1.20$ &$1.42$ &$1.56$\\
CA innovation-based $\bf Q$ & $\bf 0.55$ &$ 0.61$ &$ \bf0.97$ &$ \bf 1.08$ &$ \bf 1.30$ &$ \bf 1.45$\\
\hline
CV generative learning &$0.82$ &$0.95$ &$1.64$ &$1.97$ &$2.30$ &$2.80$\\
CA generative learning &$0.82$ &$0.96$ &$1.64$ &$1.98$ &$2.31$ &$2.83$\\
\hline
CV scaling method &$0.57$ &{$\bf 0.60$} &$1.01$ &$1.09$ &$1.35$ &$1.46$\\
CA scaling method &$0.61$ &$0.65$ &$1.09$ &$1.18$ &$1.45$ &$1.57$\\
\hline
\end{tabular}
\end{center}
\end{table*}
\subsection{Learning-Based Approaches}
{In this part, the training process of the learning algorithm given the three features (speed, road curvature, and measurement noise covariance) and its integration in the model-based KF framework is addressed.}
\subsubsection{{SL Training Procedure}}
Applying {learning-based adaptive approaches} can be made after they are appropriately trained. Several regression and classification models were trained and analyzed for both CV and CA datasets (Section {3.4}) to determine which one of them to employ in our proposed approach. The motivation to include also the classification models, aside from the regression models, is their higher robustness as there exist a finite set of accessible values for $q$. The training/test procedure included the five-fold cross-validation approach \cite{refaeilzadeh2009cross} for obtaining robust trained models. \\
The regression models included decision trees, linear regression, Gaussian process regression (GPR) \cite{seeger2004gaussian}, and support vector machine regression \cite{chapelle1999model}. Classification models were also trained, including decision trees, k nearest neighborhood (KNN), SVM, and Naive Bayes. The training and testing were made on the dataset as defined in Section {3.4} It was observed that an optimized GPR, with a non-isotropic Matern 5/2 kernel, obtained the lowest RMSE of $0.67[m]$ and thus was selected as a possible candidate from all regression models for the CV dataset. Also, SVM with a cubic kernel was selected for the classification approach, with an accuracy of $72\%$. The SVM confusion matrix is presented in Figure 12, showing that most of the values were predicted correctly. For the CA dataset, a fine tree was selected as a possible candidate as it obtained the lowest RMSE of $0.83[m]$ out of all regression models. Also, SVM with a quadratic kernel was selected for the classification approach, as it obtained an accuracy of $74.5\%$. The SVM confusion matrix is given in Figure 13, showing that most of the values were predicted correctly.

\begin{figure}[h]
\centering
{\includegraphics[width=0.48\textwidth]{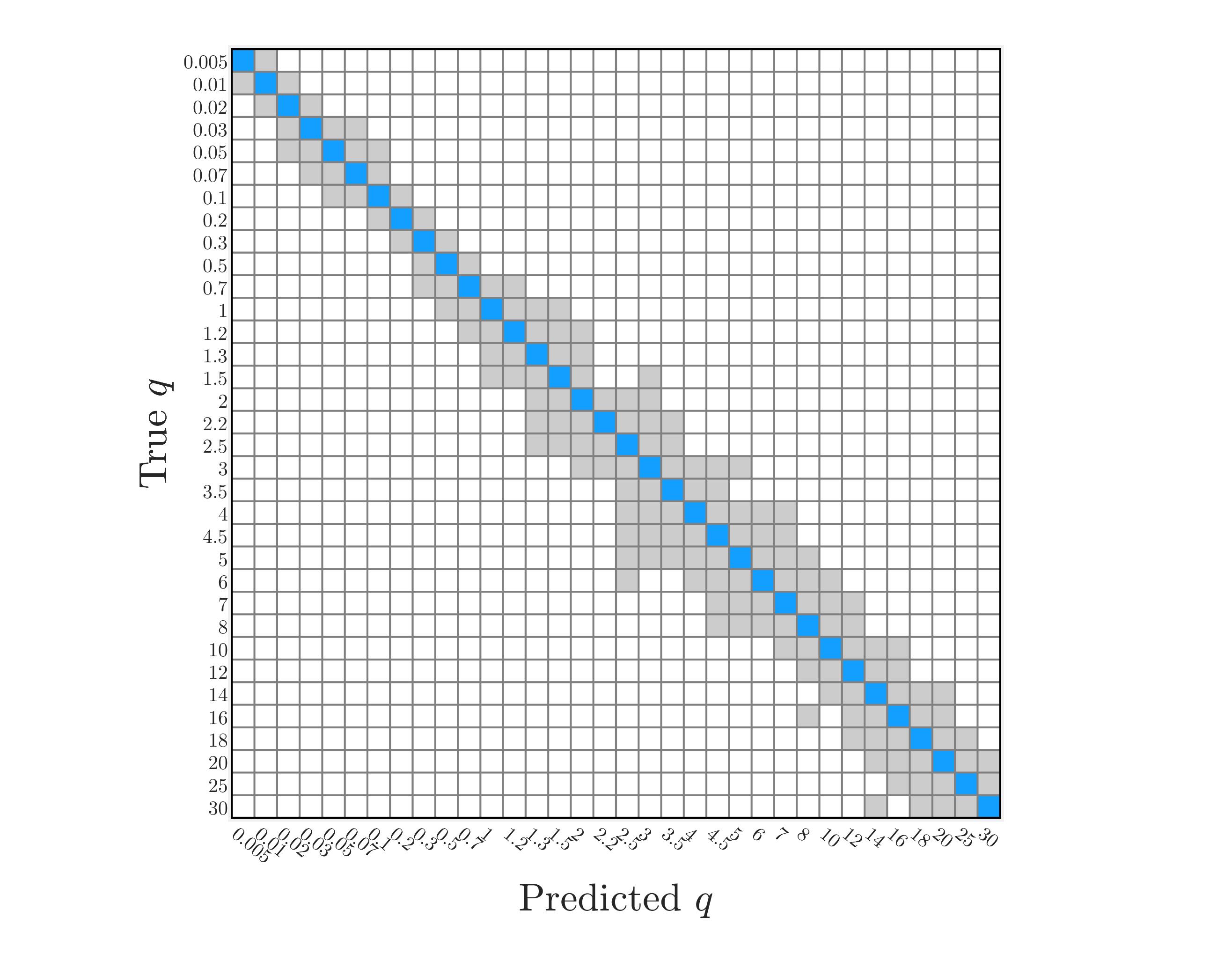}\label{fig11}}
\caption{Confusion matrix for SVM classifier with a cubic kernel, where the problem is formulated as a classification task (CV dataset).}
\end{figure}

\begin{figure}[h]
\centering
{\includegraphics[width=0.48\textwidth]{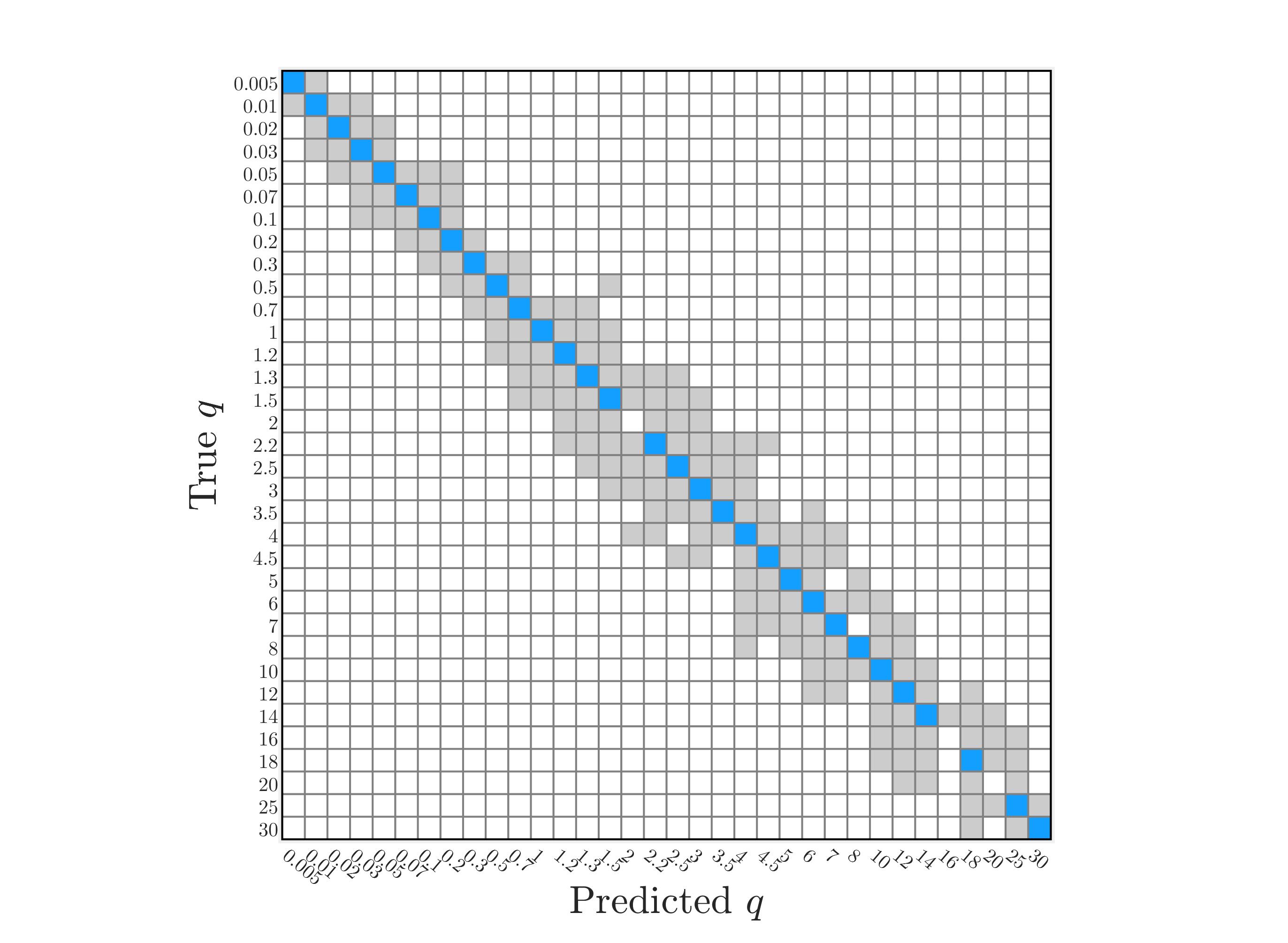}\label{fig12}}
\caption{Confusion matrix for the SVM classifier with a quadratic kernel, where the problem is formulated as a classification task (CA dataset).}
\end{figure}
\subsubsection{Integrating the ML models in the KF}
Eventually, these ML models $are$ integrated in the KF in a real-time manner to predict and tune the sub-optimal ${\bf Q}_k$ matrix {and, thus, creating an hybrid learning algorithm. To that end,} the GPR, SVM, and fine tree methods, for both CV and CA models were evaluated for two different cases: 
\begin{itemize}
\item
Using only two features: the measurement noise covariance ($r$) and speed ($s$). In that manner, the Bi-LSTM network is not required in the process.
\item
The same two features from the first case with the addition of the curvature ($\kappa$) feature. In this case, the Bi-LSTM based model was plugged into the scheme, as shown in Figure 5.
\end{itemize}
Similarly to the model-based models {(showing results in Table 3), the RobotCar dataset was used as the test dataset to evaluate the ML-based approaches}. {The learning-based approach} results are summarized in Table {4}. \\
The minimum PRMSE and PMAE for $r=0.5[m^2]$ was obtained for the CA model with the fine tree and Bi-LSTM regression models {(three features)}. Here, a PRMSE of $0.47[m]$ was {obtained, reflecting a reduction} of {$13\%$} from the CV GPR model (PRMSE=$0.54[m]$). Hence, for a low covariance measurement noise of $r=0.5[m^2]$, adding the curvature as a feature for determining ${\bf Q}_k$ improves the filter's performance. 
\\
In a higher measurement noise covariance, $r=2[m^2]$, {the results degrade and yield an} PRMSE= $0.85[m]$ and PMAE=$0.95$ for a CA model with the SVM approach (only $(s,r)$ as features). The reason for not considering the curvature lies in the difficulty of reconstructing the averaged curvature from a noisy series. {Still,} those results are lower by {$12\%$} than the one obtained using the {innovation-based} method with the CA model (PRMSE=$0.97[m]$).

Lastly, increasing the measurement noise covariance to $r=4[m^2]$ results in a lower PRMSE =$1.11[m]$ and PMAE=$1.25[m]$ for a CV model with the GPR approach {(a PMAE of 1.25[m] was obtained also in the CV GPR BiLSTM model)}. This PRMSE is lower by {$14.6\%$} than the one achieved using the CA innovation-based $\bf Q$ approach. We {also} considered the case of perfect curvature, $\kappa_{GT}$, to evaluate the curvature-based models performance. The results are also summarized in Table {4}, where there is a minor improvement of no more than $2\%$ PRMSE reduction while the $\kappa_{GT}$ is considered. Hence, the Bi-LSTM based curvature learning was well trained and provides an accurate curvature in a real time setting.

\begin{table*}[h!]
\caption {{learning-based adaptive approaches comparison for six learning-based approaches on the test dataset with different values of the measurement noise covariance. Each result is a mean of 20 repetitions and each approach was examined on the CV and CA models.}}  \label{tab:title} 
\begin{center}
\begin{tabular}{ |c|c|c|c|c|c|c| } 
\hline
Model / Error metric [m] & PRMSE & PMAE & PRMSE & PMAE & PRMSE & PMAE \\
\hline
Measurement noise covariance & $r=0.5[m^2]$ & $r=0.5[m^2]$ & $r=2[m^2]$ & $r=2[m^2]$ & $r=4[m^2]$ & $r=4[m^2]$ \\
\hline
CV GPR ${\bf Q}(r,s)$ Reg. &$0.54$ &$0.57$ & $0.86$&$0.97$ &$\bf 1.11$&$\bf 1.25$\\ 
CV SVM ${\bf Q}(r,s)$ Class. &$0.63$ &$0.71$ &$0.94$ &$1.05$ &$1.16$ &$1.30$ \\
\hline
CA fine tree ${\bf Q}(r,s)$ Reg.  & $0.62$ &$0.70$ &$1.13$ &$1.27$ &$1.52$ &$1.72$\\
CA SVM ${\bf Q}(r,s)$ Class.  &$0.63$ &$0.70$ &$\bf 0.85$ &$\bf 0.95$ &$1.15$ &$1.29$\\
\hline
CV GPR BiLSTM {${\bf Q}(r,s,\hat \kappa)$} Reg. &{$0.49$} &{$0.54$} &{$0.86$} &{$\bf 0.95$} &{$1.14$} &{$\bf 1.25$}\\ 
CV SVM BiLSTM {${\bf Q}(r,s,\hat \kappa)$} Class. &{$0.51$} &{$0.55$}&{$0.91$} &{$0.99$} &{$1.22$} &{$1.32$} \\
\hline
CA fine tree BiLSTM {${\bf Q}(r,s,\hat \kappa)$} Reg.  &{$\bf 0.47$} &{$\bf 0.53$} &{$\bf 0.85$} &{$\bf 0.95$} &{$1.14$} &{$1.27$}\\
CA SVM BiLSTM {${\bf Q}(r,s,\hat \kappa)$} Class.  &{$0.50$} &{$0.56$} &{$0.90$} & {$0.99$} &{$1.20$} &{$1.33$}\\
\hline
{CV GPR BiLSTM ${\bf Q}(r,s,\kappa_{GT})$ Reg. }&{$0.49$} &{$0.54$} &{$0.86$} &{$0.96$} &{$1.15$} &{$1.27$}\\ 
{CV SVM BiLSTM ${\bf Q}(r,s,\kappa_{GT})$ Class.} &{$0.50$} &{$0.54$}& {$0.98$} &{$1.04$} &$1.30$ &{$1.38$} \\
\hline
{CA fine tree BiLSTM ${\bf Q}(r,s,\kappa_{GT})$ Reg. } &{$ 0.48$} &{$ 0.53$} &{$0.86$} &{$0.96$} &{$1.15$} &{$1.29$}\\
{CA SVM BiLSTM ${\bf Q}(r,s,\kappa_{GT})$ Class.}  &{$0.50$} &{$0.56$} &{$0.90$} &{$1.00$} &{$1.21$} &{$1.34$}\\
\hline
\end{tabular}
\end{center}
\end{table*}
\subsection{{Discussion}}
The {model-based adaptive approaches} results are provided in Section {4.1} and our hybrid {learning-based adaptive approaches} results in Section {4.2}.  
The performance of the adaptive model-based approaches is superior to model-based approaches with constant process noise values, as they try to capture the vehicle current dynamics and the measurement noise statistics in a real-time {setting}. The innovation-based model and the scaling method are better than the generative learning, as they achieved lower PRMSE and PMAE for all tested measurement noise covariance values. For $r=0.5[m^2]$, the CA innovation-based $\bf Q$ obtained PRMSE of $0.55[m]$ and PMAE of $0.61[m]$. Thus, an improvements of $11\%$ and $13\%$ was obtained in the PRMSE and PMAE, respectively, compared to model-based approaches with constant process noise values.\\
The {learning-based adaptive approaches} were pre-trained to capture the non-linearity and other uncertainty properties; once using the relationship between the measurement noise covariance and vehicle speed and once with the addition of the road curvature as a feature. The hybrid {learning-based adaptive approaches} achieved better performance than the adaptive models as they consider additional learned information during the training phase. 
Without the curvature feature, CV GPR-based ${\bf Q}(r,s)$ Reg. model, obtained a PRMSE of $0.54[m]$ and PMAE of $0.57[m]$, for $r=0.5[m^2]$. This model improves the PRMSE by $13\%$ and the PMAE by $19\%$ compared to model-based approaches with constant process noise values. Including the road curvature feature, the best performance for the CA model was archived with the fine tree BiLSTM-based ${\bf Q}(r,s,\hat \kappa)$. It obtained a PRMSE of {$0.47[m]$} and PMAE of {$0.53[m]$}, for $r=0.5[m^2]$. Hence, it improves the PRMSE by {$24\%$} and the PMAE by {$24\%$} compared to model-based approaches with the CA constant process noise values and improves the PRMSE with {$14.5\%$} and the PMAE with {$13.1\%$} compared to the adaptive model-based methods. 

\section{Conclusions}
The proper choice of the trajectory process noise covariance matrix is critical for achieving high accuracy when implementing the discrete {linear} KF for vehicle tracking scenarios. The commonly used solution is to choose constant matrix parameters or to use any model-based adaptive approach. A practical {hybrid} ML-based framework to tune a sub-optimal covariance matrix online, was proposed. The framework combines kinematic and geometrical features, with a Bi-LSTM-based model to estimate the road curvature and an SL model utilizing it with the measurement noise covariance and vehicle speed {to regress the process noise matrix.}

{The proposed scheme for vehicle tracking was thoroughly evaluated using extensive simulations on the RobotCar dataset. The results demonstrate the efficiency of the suggested approach as it outperforms model-based adaptive Kalman filter approaches in terms of the position RMSE. The proposed approach requires training the model only once, regardless of the specific characteristics of the vehicle's motion or the road's trajectory.\\
Specifically, our learning approach achieved a position RMSE lower than $0.5[m]$ ($24\%$ improvement compared to the best model-based approach) using a Kalman filter with a CA model, by incorporating Bi-LSTM for the road curvature regression and SVM for the process noise regression. The SVM algorithm uses only three features: vehicle speed, measurement noise covariance, and road curvature. These results demonstrate the effectiveness of the proposed method in handling the complexities of real-world vehicle tracking and pave the way for future research in this field.} The proposed approach can be implemented in real-time vehicle tracking problems {( for example using a standalone GNSS receiver)}  including autonomous vehicles, robotics platforms, or drones.\\
{The proposed learning approach requires additional computational resources when compared to existing model-based adaptive approaches. Thus, an increase in the computational load of the system is expected. However, with the advancement in hardware and optimization techniques, this drawback can be mitigated. In future research, we aim to elaborate the proposed learning approaches by evaluating it on additional experimental data to ensure their robustness. }

\appendix
\section{List of $q$ candidates}
${\cal Q}=\left\{ \begin{array}{l}
0.005,0.01,0.02,0.03,0.05,0.07,0.1,0.2,\\
0.3,0.5,0.7,1,1.2,1.3,1.5,2,2.5,3,3.5,4,\\
4.5,5,6,7,8,10,12,14,16,18,20,25,30
\end{array} \right\}$

\bibliographystyle{unsrt}
\bibliography{cas-refs}

\end{document}